\documentclass{SCIS2024}
\usepackage{makecell}
\usepackage{multirow}
\usepackage{multicol}
\usepackage{adjustbox}
\usepackage{wrapfig}
\usepackage{picins}
\usepackage{float}

\begin{document}
%\oa
%%%%%%%%%%%%%%%%%%%%%%%%%%%%%%%%%%%%%%%%%%%%%%%%%%%%%%%
%%% Authors do not modify the information below
%%% ×÷Õß²»ÐèÒªÐÞ¸Ä´Ë´¦ÐÅÏ¢
\ArticleType{REVIEW}
%\SpecialTopic{}
%\luntan
\Year{2024}
\Month{}
\Vol{}
\No{}
\DOI{}
\ArtNo{}
\ReceiveDate{}
\ReviseDate{}
\AcceptDate{}
\OnlineDate{}
%%%%%%%%%%%%%%%%%%%%%%%%%%%%%%%%%%%%%%%%%%%%%%%%%%%%%%%

%%% title: ±êÌâ
\title{An overview of domain-specific foundation model: key technologies, applications and challenges}{}

%%% Corresponding author: Í¨ÐÅ×÷Õß
%%%   \author[number]{Full name}{{email@xxx.com}}
%%% General author: Ò»°ã×÷Õß
%%%   \author[number]{Full name}{}
\author[1,2]{Haolong CHEN}{}
\author[1,10]{Hanzhi CHEN}{}
\author[1,6]{Zijian ZHAO}{}
\author[4]{Kaifeng HAN}{{hankaifeng@caict.ac.cn}}
\author[1,2]{\\Guangxu ZHU}{{gxzhu@sribd.cn}}
\author[5]{Yichen ZHAO}{}
\author[3,4]{Ying DU}{}
\author[7,8]{Wei XU}{}
\author[1,9]{Qingjiang SHI}{}

%%% Author information for page head. Ò³Ã¼ÖÐµÄ×÷ÕßÐÅÏ¢
\AuthorMark{Chen H L}

%%% Authors for citation. Ê×Ò³ÒýÓÃÖÐµÄ×÷ÕßÐÅÏ¢
\AuthorCitation{Chen H L, Chen H Z, Zhao Z J, et al}

%%% Authors' contribution. Í¬µÈ¹±Ï×
%\contributions{Authors A and B have the same contribution to this work.}

%%% Address. µØÖ·
%%%   \address[number]{Affiliation, City {\rm Postcode}, Country}

\address[1]{Shenzhen Research Institute of Big Data, Shenzhen {\rm 518172}, China}
\address[2]{School of Science and Engineering, The Chinese University of Hong Kong, Shenzhen {\rm 518172}, China}
\address[3]{Department of Electronic Engineering and Information Science, University of Science and Technology of China, \\Hefei {\rm 230026}, China}
\address[4]{The Mobile Communications Innovation Center, China Academy of Information and Communications Technology, \\Beijing {\rm 100191}, China}
\address[5]{China Mobile Group Device Co., Ltd., Beijing {\rm 100033}, China}
\address[6]{School of Computer Science and Engineering, Sun Yat-sen University, Guangzhou {\rm 510275}, China}
\address[7]{National Mobile Communications Research Laboratory, Southeast University, Nanjing {\rm 210096}, China}
\address[8]{Purple Mountain Laboratories, Nanjing {\rm 211111}, China}
\address[9]{School of Software Engineering, Tongji University, Shanghai {\rm 201804}, China}
\address[10]{School of Electrical and Electronic Engineering, Nanyang Technological University, Singapore {\rm 639798}, Singapore}

%%% Abstract. ÕªÒª
\abstract{The impressive performance of ChatGPT and other foundation-model-based products in human language understanding has prompted both academia and industry to explore how these models can be tailored for specific industries and application scenarios. This process, known as the customization of domain-specific foundation models (FMs), addresses the limitations of general-purpose models, which may not fully capture the unique patterns and requirements of domain-specific data. Despite its importance, there is a notable lack of comprehensive overview papers on building domain-specific FMs, while numerous resources exist for general-purpose models. To bridge this gap, this article provides a timely and thorough overview of the methodology for customizing domain-specific FMs. It introduces basic concepts, outlines the general architecture, and surveys key methods for constructing domain-specific models. Furthermore, the article discusses various domains that can benefit from these specialized models and highlights the challenges ahead. Through this overview, we aim to offer valuable guidance and reference for researchers and practitioners from diverse fields to develop their own customized FMs.}

%%% Keywords. ¹Ø¼ü´Ê

\maketitle

%%%%%%%%%%%%%%%%%%%%%%%%%%%%%%%%%%%%%%%%%%%%%%%%%%%%%%%
%%% The main text. ÕýÎÄ²¿·Ö
%%%%%%%%%%%%%%%%%%%%%%%%%%%%%%%%%%%%%%%%%%%%%%%%%%%%%%%
\section{Introduction} \label{sec:introduction}

\begin{figure}[!t]
    \centering
    \centerline{\includegraphics[width=1.0\linewidth]{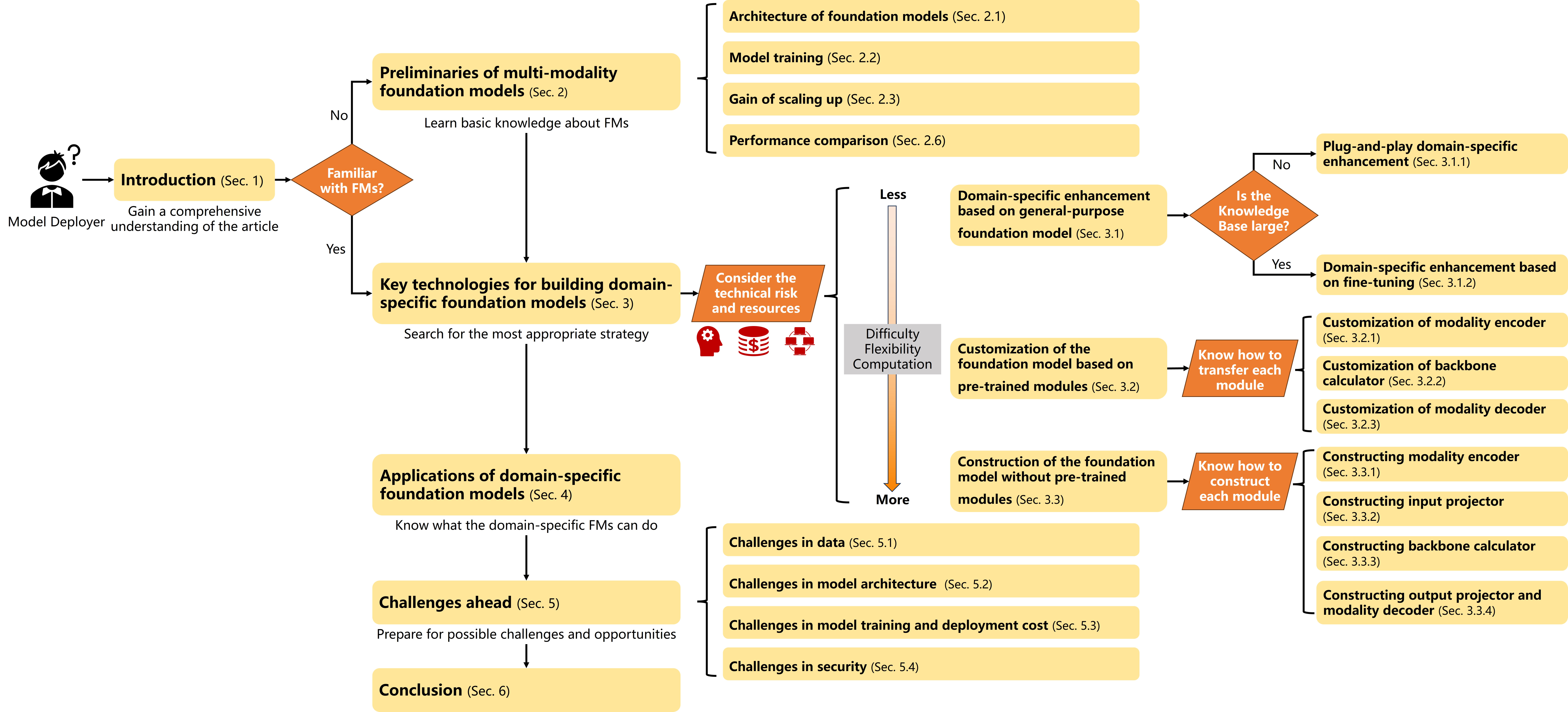}}
    \caption{{The organization structure of this article.}}
    \label{overall_framework}
\end{figure}

ChatGPT has redefined people's understanding of artificial intelligence with its outstanding performance. As its core technology, the Large Language Model (LLM) has become an essential tool for researchers and practitioners across various fields to improve their workflows. General-purpose foundation models (FMs) are usually trained on large public datasets, enabling them to learn and address a wide range of common problems. However, these datasets cannot fully encompass all the specialized knowledge and technical details of specific domains. As a result, although general-purpose FMs possess broad general knowledge, they lack the necessary depth to meet the complex needs of some specific fields~\cite{ding2024survey}. Therefore, constructing domain-specific FMs tailored to the needs of particular industries has become particularly important. The term ``domain" refers to a specific area of knowledge with logical completeness and knowledge complexity, such as healthcare, finance, or legal services. Often, only professionally trained one can master a certain domain's terminology, concepts, and techniques. Domain-specific FMs, or industry-specific FMs, are developed using data specific to a particular field. Compared to general-purpose FMs, they are trained with a large amount of domain-specific data, enabling them to more accurately understand the unique language, terminology, concepts, and nuances inherent to the specific domain and generate domain-specific professional content.

With the widespread use of ChatGPT-like products, the scope of the ``foundation model" is gradually expanding. It is necessary to clearly define the foundation model discussed to support the subsequent discussion on customizing domain-specific FMs. The FMs mentioned in this article are neural network models that consist of at least one of the five modules (which will be detailed later) in a general multi-modality FM. These models also exhibit the following characteristics:

\begin{itemize}
    \item \textbf{Big data}: The model utilizes a large volume of data covering various scenarios for training to provide ample knowledge for the model.
    \item \textbf{Large parameters}: The model possesses a huge number of parameters sufficient to embed the knowledge implied by the big data into the model's parameters.
    \item \textbf{Versatility}: The model's input data format and data processing workflow can adapt to different requirements across various task scenarios.
    \item \textbf{Fast adaptation capabilities}: The model possesses a vast knowledge repository, enabling it to deliver robust performance even in unknown data domains with minimal fine-tuning.
\end{itemize}

In the context of artificial intelligence, we use ``modality" to categorize data with a consistent data structure and semantic structure. Such a definition brings the conceptual consistency of each modality, which is the theoretical foundation of multi-modality data analysis and processing. Based on the number of modalities an FM can process, they can be classified into single-modality FMs and multi-modality FMs, as shown in Table~\ref{tab_foundation_models}. With the certain modality an FM can process, we can use different names to categorize FMs, such as large language models, large vision models, large vision-language models, etc.

\begin{table*}[t!]
    \footnotesize
    \caption{Examples of FMs with two different types}
    \label{tab_foundation_models}
    \def\tabblank{\hspace*{10mm}}
    
    \begin{tabularx}{\textwidth}{@{\extracolsep{\fill}}c>{\centering\arraybackslash}p{11cm}}
    \toprule
    Types of FMs & Examples \\
    \hline
    \multirow{3}*{Single-modality FMs} & VGG~\cite{simonyan2014very}, ResNet~\cite{he2016deep}, GPT-1~\cite{radford2018improving}, GPT-2~\cite{radford2019language}, GPT-3~\cite{brown2020language}, GPT-3.5 turbo, BERT~\cite{devlin2018bert}, GLM~\cite{du2021glm, zeng2022glm}, LLaMA~\cite{touvron2023llama}, LLaMA-2~\cite{touvron2023llama2}, iGPT~\cite{chen2020generative}, LVM~\cite{bai2024sequential}, SAM\cite{kirillov2023segment}, BART~\cite{lewis2019bart}, T5~\cite{raffel2020exploring}, Time-LLM~\cite{jin2023time}, UniTS~\cite{gao2024units}, ST-LLM~\cite{liu2024spatial} \\
    \hline
    \multirow{2}*{Multi-modality FMs} & CoDi~\cite{tang2024any}, CoDi-2~\cite{tang2024codi}, Claude-3, GPT-4~\cite{achiam2023gpt}, LLaVA~\cite{liu2024visual}, BriVL~\cite{fei2022towards}, ImageBind~\cite{girdhar2023imagebind}, NExT-GPT~\cite{wu2023next} \\
    \bottomrule
    \end{tabularx}
\end{table*}

In constructing domain-specific FMs, a series of challenges will arise, especially in the data acquisition and preprocessing stages. For example, the domain-specific data required may not be open-source or readily accessible, as it often has high confidentiality. Additionally, the modalities of domain-specific data might differ from those used for training general-purpose FMs, making it difficult to adapt existing models to process this data. Furthermore, the environments where the domain-specific data is collected may differ significantly from those for the pre-training datasets, resulting in domain-specific knowledge that pre-trained models are unfamiliar with. In general, constructing a domain-specific FM is challenging and costly, with significant implications for technical security, but it is expected to yield high economic benefits. Therefore, it is essential to thoroughly review and explore the methodologies for constructing these models, providing guidelines for both researchers and practitioners. The organization structure of this article is summarized in Fig.~\ref{overall_framework}.

Notably, previous review articles have predominantly focused on developing general-purpose FMs. Recently, while several review articles have begun to explore the domain-specific adaptation of FMs, there is a significant gap in the literature for a comprehensive exploration of adaptation strategies that apply to FMs of all modalities, extending beyond language, vision, or any single modality, to various application domains. We summarize representative surveys or review articles on FMs in Table~\ref{Related_Survey}. This paper aims to provide researchers and practitioners interested in building domain-specific FM applications with methodological references by introducing the key methodologies. Also, practical examples and future directions will be discussed.

\begin{table*}[!t]
    \footnotesize
    \caption{Summary of representative surveys or review articles on FMs}
    \label{Related_Survey}
    \def\tabblank{\hspace*{10mm}}
    
    \begin{tabularx}{\textwidth}{@{\extracolsep{\fill}}c>{\centering\arraybackslash}p{4cm}ccc}
    \toprule
    Article & Modality & Domain Adaptation & Domain & Year \\ \hline
    Zhao \textit{et al.}~\cite{zhao2023survey} & Language & No & General & 2023 \\
    Liu \textit{et al.}~\cite{liu2023towards} & Graph & No & General & 2023 \\
    Mao \textit{et al.}~\cite{mao2024graph} & Graph & No & General & 2024 \\
    Du \textit{et al.}~\cite{du2022survey} & Vision-Language & No & General & 2022 \\
    Yin \textit{et al.}~\cite{yin2023survey} & Multi-modality-Language & No & General & 2023 \\
    Yeh \textit{et al.}~\cite{yeh2023toward} & Time-series Data & No & General & 2023 \\
    \multirow{2}*{Jin \textit{et al.}~\cite{jin2023large}} & Time-series \& Spatio-temporal Data & \multirow{2}*{No} & \multirow{2}*{General} & \multirow{2}*{2023} \\
    Cao \textit{et al.}~\cite{cao2023comprehensive} & Multi-modality & No & General & 2023 \\
    Zhou \textit{et al.}~\cite{zhou2023comprehensive} & Multi-modality & No & General & 2023 \\
    Zhang \textit{et al.}~\cite{zhang2024mm} & Multi-modality-Language & No & General & 2024 \\
    \hline
    Lai \textit{et al.}~\cite{lai2023large} & Language & Yes & Law & 2024 \\
    Ahn \textit{et al.}~\cite{ahn2024large} & Language & Yes & Mathematics & 2024 \\
    Li \textit{et al.}~\cite{li2023large} & Language & Yes & Finance & 2023 \\
    Thirunavukarasu \textit{et al.}~\cite{thirunavukarasu2023large} & Language & Yes & Medicine & 2023 \\
    Kazerouni \textit{et al.}~\cite{kazerouni2023diffusion} & Vision & Yes & Medicine & 2023 \\
    Shahriar~\cite{shahriar2022gan} & Multi-modality & Yes & Arts & 2022 \\
    Hou \textit{et al.}~\cite{hou2023large} & Language & Yes & Software Engineering & 2023 \\
    Bariah \textit{et al.}~\cite{bariah2023large} & Language & Yes & Telecommunications & 2024 \\
    Zhou \textit{et al.}~\cite{zhou2024large} & Language & Yes & Telecommunications & 2024 \\
    Cui \textit{et al.}~\cite{cui2024survey} & Multi-modality & Yes & Autonomous Driving & 2024 \\
    Yang \textit{et al.}~\cite{yang2024harnessing} & Language & Yes & General & 2024 \\
    \hline
    Ours & Multi-modality & Yes & General & - \\
    \bottomrule
    \end{tabularx}
\end{table*}

\section{Preliminaries of multi-modality foundation models}
This section will elaborate on the preliminary technological basis for customizing domain-specific FMs. We begin with the architecture of FMs, detailing all functional modules. Then, from two perspectives - model training and scaling gain - we will explain the foundational technologies that enable each module to contribute to the high performance of FMs. Finally, we compare the performance of various general-purpose large multi-modality models for an intuitive sense of the performance that FMs can achieve and guide model employers in choosing the most suitable FM.

\subsection{Architecture of foundation models}\label{Architecture of FMs}
According to state-of-the-art research on FMs, it is widely considered that multi-modality FMs can encompass all functionalities and structures of single-modality FMs. Essentially, a single-modality FM implements only a subset of the functionalities of a multi-modality FM.

\begin{figure}[!t]
    \centering
    \centerline{\includegraphics[width=1.0\linewidth]{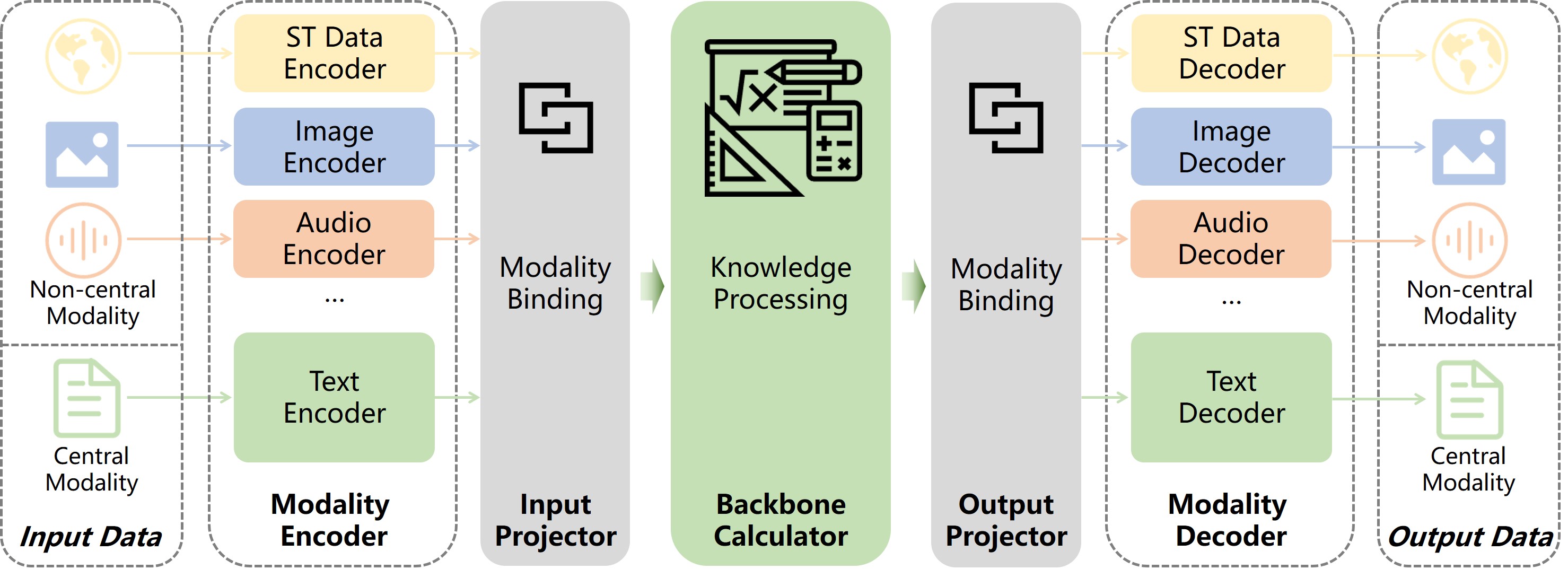}}
    \caption{The framework of multi-modality FMs with language as the central modality.}
    \label{fig1}
\end{figure}

The five-module framework proposed for multi-modality FMs in the paper~\cite{zhang2024mm} effectively encompasses the architecture of multi-modality FMs with language as the central modality. However, the recent emergence of non-language FMs, including vision FMs~\cite{bai2024sequential, kirillov2023segment}, graph FMs~\cite{liu2023towards, mao2024graph}, time series FMs and spatio-temporal (ST) FMs~\cite{yeh2023toward, jin2023large}, indicates that the backbone of FMs is expanding beyond language modalities. Therefore, we propose that the structure of multi-modality FMs can be divided into the following five modules: Modality Encoder, Input Projector, Backbone Calculator, Output Projector, and Modality Decoder. Figure~\ref{fig1} illustrates the framework of a multi-modality FM with language as the central modality.

For multi-modality FMs, we define the set of all input modalities as $M$. Generally, a multi-modality FM has a central modality $C$. Using modality alignment techniques, the multi-modality FM projects all the modalities it can handle onto this central modality. Below, we define the five modules of the multi-modality FM and the input and output data for each module, laying down the foundation for describing the architecture of FMs in this paper.

\textbf{Modality Encoder} (ME) encodes the data $D_X$ of an input modality $X$ into a feature vector $F_X$:
\begin{equation}
F_{X} = \text{ME}_X(D_{X}), \quad {X} \in {M}.
\end{equation}

% The goal of ME is to minimize the difference between the output feature and the groundtruth feature of modality $C$.

\textbf{Input Projector} (IP) projects the feature vector $F_{X}$ of a modality $X$ into the feature vector $F_{C}$ of the central modality $C$:
\begin{equation}
F_{C} = \text{IP}_{XC}(F_{X}), \quad {X},{C} \in {M}.
\end{equation}

\textbf{Backbone Calculator} (BC) performs operations on the feature vector $F_{C}$ of the central modality $C$, yielding results such as inference and generation $\hat{F_{C}}$:
\begin{equation}
\hat{F_{C}} = \text{BC}_C(F_{C}), \quad {C} \in {M}.
\end{equation}

\textbf{Output Projector} (OP) projects the feature vector $\hat{F_{C}}$ of the central modality $C$ into the feature vector $\hat{F_{X}}$ of a modality $X$:
\begin{equation}
\hat{F_{X}} = \text{OP}_{CX}(\hat{F_{C}}), \quad {X},{C} \in {M}.
\end{equation}

\textbf{Modality Decoder} (MD) decodes the feature vector $\hat{F_{X}}$ of the output modality $X$ back into the original data format, resulting in the decoded data $\hat{D_{X}}$:
\begin{equation}
\hat{D_{X}} = \text{MD}_X(\hat{F_{X}}), \quad {X} \in {M}.
\end{equation}

According to the above definition, building a domain-specific FM entails choosing the necessary modules --- some of which may be optional --- and assembling a model tailored to the specific domain requirements, followed by training it with data pertinent to that domain.

Here, we provide a concrete example for better illustration. CoDi-2~\cite{tang2024codi} is a language-centric multi-modality FM capable of handling various modalities such as images, videos, text, and audio. It utilizes a large language model (LLM) to generate new content. CoDi-2 employs Vision Transformers (ViTs) pre-trained by ImageBind~\cite{girdhar2023imagebind} as the MEs for the image, video, and audio modalities. For the text modality, the ME is a text tokenizer. Since the output features of the ImageBind pre-trained ViT are specific to the image modality, CoDi-2 trains a multilayer perceptron (MLP) as the IP to map the features from the Image, Video, and Audio MEs to the central modality, Text. Subsequently, these Text features are fed into its BC, LLaMA2—an LLM—that generates content matching the user's input prompt in the Text modality. Then, the generated Image and Audio content are mapped back to their respective feature spaces through an OP, another MLP. Finally, pre-trained diffusion models are used as Image, Video, and Audio MDs to generate the content.

\subsection{Model training} \label{Model Training}

After constructing an FM module by module, we need to train it. Model training involves setting up various training tasks to enable FM to specifically extract the features required for solving each task. Feature extraction concerns extracting representative features from raw data. As raw data often contains a significant amount of redundant and noisy information, feature extraction, which maps the data into a more information-dense feature space, enables the models to understand data structure and patterns more effectively. Feature space is the set of coordinates representing data samples across multiple feature dimensions, which mathematically refer to a high-dimensional space. Note that ``raw data'' and ``feature" are used in this paper to differentiate whether a dataset has undergone feature extraction, which is essentially a set of mathematical calculations that increase the information density of the data. In deep learning, neural networks can perform end-to-end feature extraction from raw data. However, this often requires large amounts of data and computational resources to ensure good performance and generalization. In each layer of neural networks, it transfers the output of the previous layer to a new vector space. This structure allows for a flexible definition of the output dimensions of each layer without explicitly specifying the transformation.

\begin{wrapfigure}{r}{8cm}
    \centering
    \includegraphics[width=0.5\textwidth]{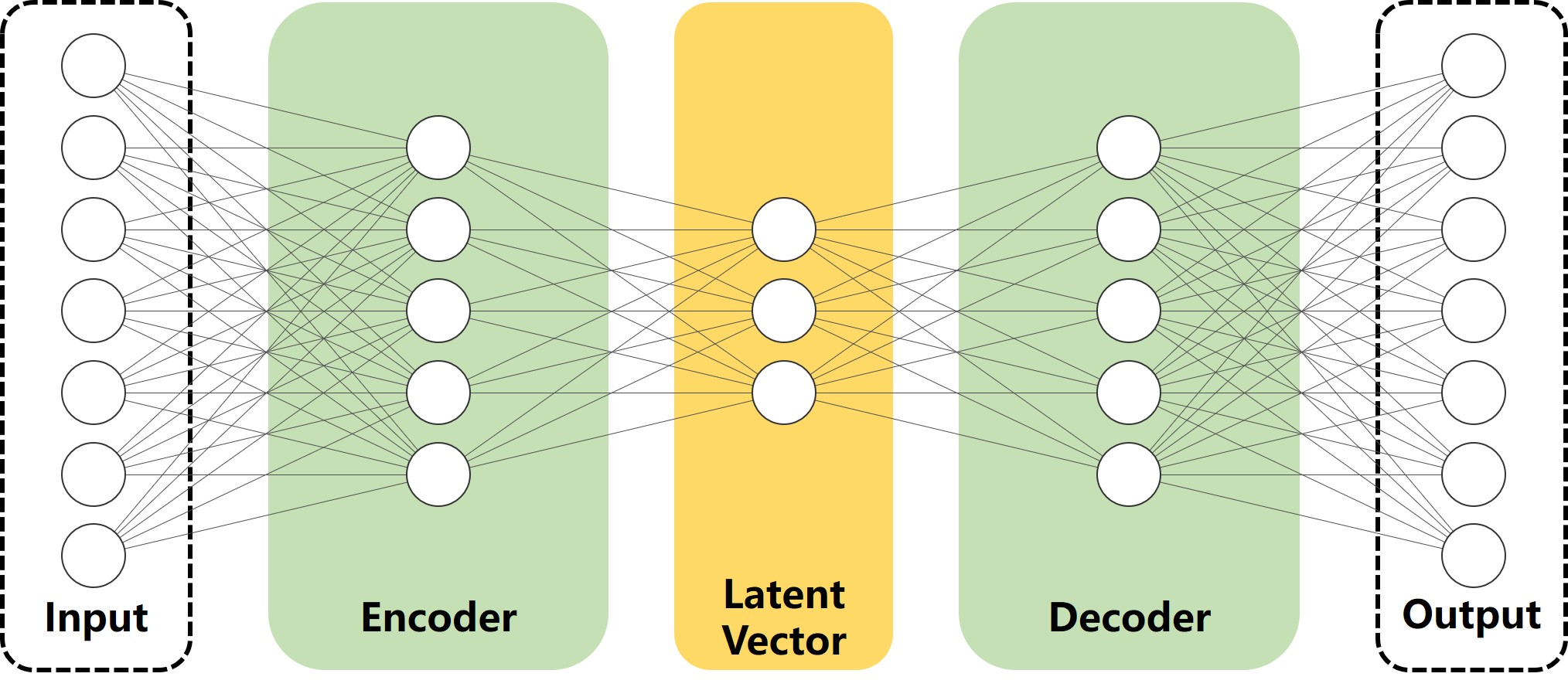}
    \caption{The structure of autoencoder.}
    \label{fig2}
\end{wrapfigure}
% 各种训练方式

Numerous training methods are available to achieve better feature extraction. They construct different loss functions to teach the model how to focus on different aspects of the data, thereby extracting various feature representations. If labeled information is available as the target for a specific task during training, supervised learning can be used to train the model's feature extraction capabilities. For example, suppose we have labeled data for a classification task. In that case, we can construct a loss function such as Cross-Entropy Loss to measure the difference between the model's predicted probability distribution and the true label's probability distribution. This provides the model with a specific direction for optimization to achieve the best feature extraction for the classification task. Sometimes, we may not have labeled data for a specific task but still wish to mine features from the data to perform better on downstream tasks in the future. In such cases, unsupervised learning methods can be employed. Unsupervised training involves learning from data without the need for labeled examples. Common methods include autoencoders, which learn to compress and reconstruct input data, as shown in Fig~\ref{fig2}, capturing important features. Another approach is using generative models like Generative Adversarial Networks (GANs) or Variational Autoencoders (VAEs) to learn the distribution of the data. Self-supervised training is also considered a subset of unsupervised training~\cite{gui2024survey}, such as in training LLMs where autoregressive prediction tasks are used. By generating subsequent content based on the previous content, the model inherently captures the distribution of text sequences. Combining unsupervised and self-supervised training can form semi-supervised training, which is well-suited for situations where there is a limited amount of labeled data and a large amount of unlabeled data.

Humans can also play a significant guiding role in training the model's feature extraction capabilities. Reinforcement Learning from Human Feedback (RLHF) is a method that combines reinforcement learning with human feedback, adjusting the model based on human evaluations of its behavior to learn feature representations that align more closely with human expectations. This technique is often mentioned in the context of training LLMs. First, the model generates a series of outputs and requests a human evaluation of these outputs, such as scoring, ranking, or providing other forms of feedback. These feedback data are then used to train a reward model that assesses the quality of different outputs based on human preferences. Then, using the reward model, the initial model is fine-tuned through reinforcement learning algorithms. During this fine-tuning process, the model continuously adjusts its behavior strategy based on the feedback from the reward model to maximize cumulative rewards. Through this process, the model can better understand user preferences and needs, generating outputs more aligned with human expectations.

% 模态对齐

For a single-modality FM, cross-modality data is unavailable, so the architecture excludes IPs and OPs. In contrast, multi-modality FMs must accommodate the processing of various data modalities, including the primary and ancillary modalities. To achieve data conversion between modalities through IPs and OPs, the key is to apply modality alignment. The goal of modality alignment is to process the feature vectors of different modalities into a common feature space with the same dimension by using a loss function to characterize the correlation between feature vectors. Ideally, modality alignment should ensure that raw data of different modalities carrying the same semantic information are represented as the same point in the target feature space, facilitating cross-modality information transfer.

There are two main implementation methods for modality alignment: the fusion encoder architecture and the dual encoder architecture~\cite{du2022survey}:

\begin{enumerate}
    \item \textbf{Fusion encoder}: Fusion encoder encodes the multi-modality data by the self-attention or cross-attention mechanism from transformer~\cite{vaswani2017attention}. The attention mechanism can be represented as follows:

    \begin{equation}
       Attenton(Q,K,V) =\text{softmax}(\frac{Q^T K}{\sqrt{d_k}})V ,
    \end{equation}
    where query $Q$, key $K$, and value $V$ are all intermediate features based on the input feature, and $d_k$ refers to the dimension of $Q$ and $K$.

    \begin{minipage}{0.94\textwidth}
        \begin{wrapfigure}{r}{0.45\textwidth}
            \centering
            \includegraphics[width=0.45\textwidth]{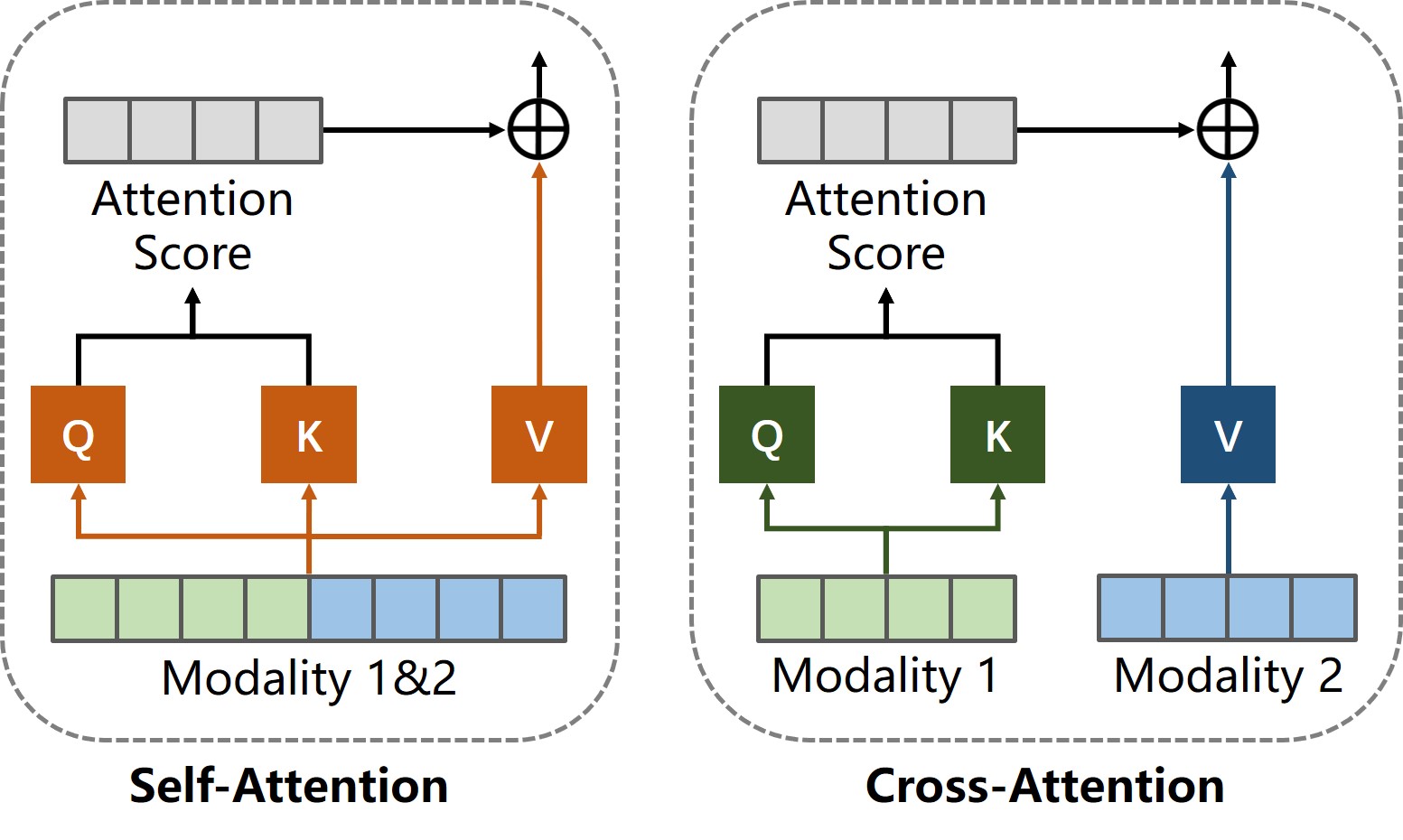}
            \caption{The structure of fusion encoder.}
            \label{fig3}
        \end{wrapfigure}
        Self-attention-based mechanisms require concatenating the feature vectors of the main and auxiliary modalities as input to the transformer to generate $Q$, $K$, and $V$, allowing the model to automatically focus on the features of different modalities and achieve cross-modality information fusion. For example, VL-BERT~\cite{su2019vl} concatenates the feature vectors of text and images, and then uses the self-attention mechanism of transformer to aggregate and align language-visual features. On the other hand, methods based on the cross-attention mechanism calculate $Q$, $K$, and $V$ separately for the feature vectors of the two modalities, thereby achieving cross-modality information fusion. For example, DiT~\cite{peebles2023scalable} captures the relationship between text and images by self-attention mechanism and then realizes the image generation controlled by text. We illustrate the two fusion encoder architectures in Figure~\ref{fig3}.
    \end{minipage}

    % \begin{figure}[!t]
    %     \centering
    %     \includegraphics[width=0.6\linewidth]{./img/3.jpg}
    %     \caption{The structure of fusion encoder.}
    %     \label{fig3}
    % \end{figure}

    % \item \textbf{Dual encoder}: As a multi-modality learning strategy, dual encoder trains a dedicated encoder for each modality independently. The core idea of this architecture is to guide the learning process of the two encoders in synchronization through semantic similarity metrics by leveraging contrastive learning methods. Then, the output feature vectors of different encoders can be projected into the same vector space. Specifically, this model is based on the hypothesis that if the feature vectors output by the two encoders belong to the same feature space, the feature vectors with paired labels should be closer in the vector space and vice versa. Through this alignment method, we can expect that the output results of encoders for different modalities describing similar objects or scenes will be close enough. And even in an ideal state, they will converge to the same point in the feature space. The key to achieving this goal is to construct a reasonable model architecture and thoroughly train it on a large-scale dataset. Figure~\ref{fig4} shows the processing logic of the dual encoder.

    \item \textbf{Dual encoder}: As a multi-modality learning strategy, dual encoder trains a dedicated encoder for
    
    \begin{minipage}{0.94\textwidth}
        \begin{wrapfigure}{r}{0.4\textwidth}
            \centering
            \includegraphics[width=0.4\textwidth]{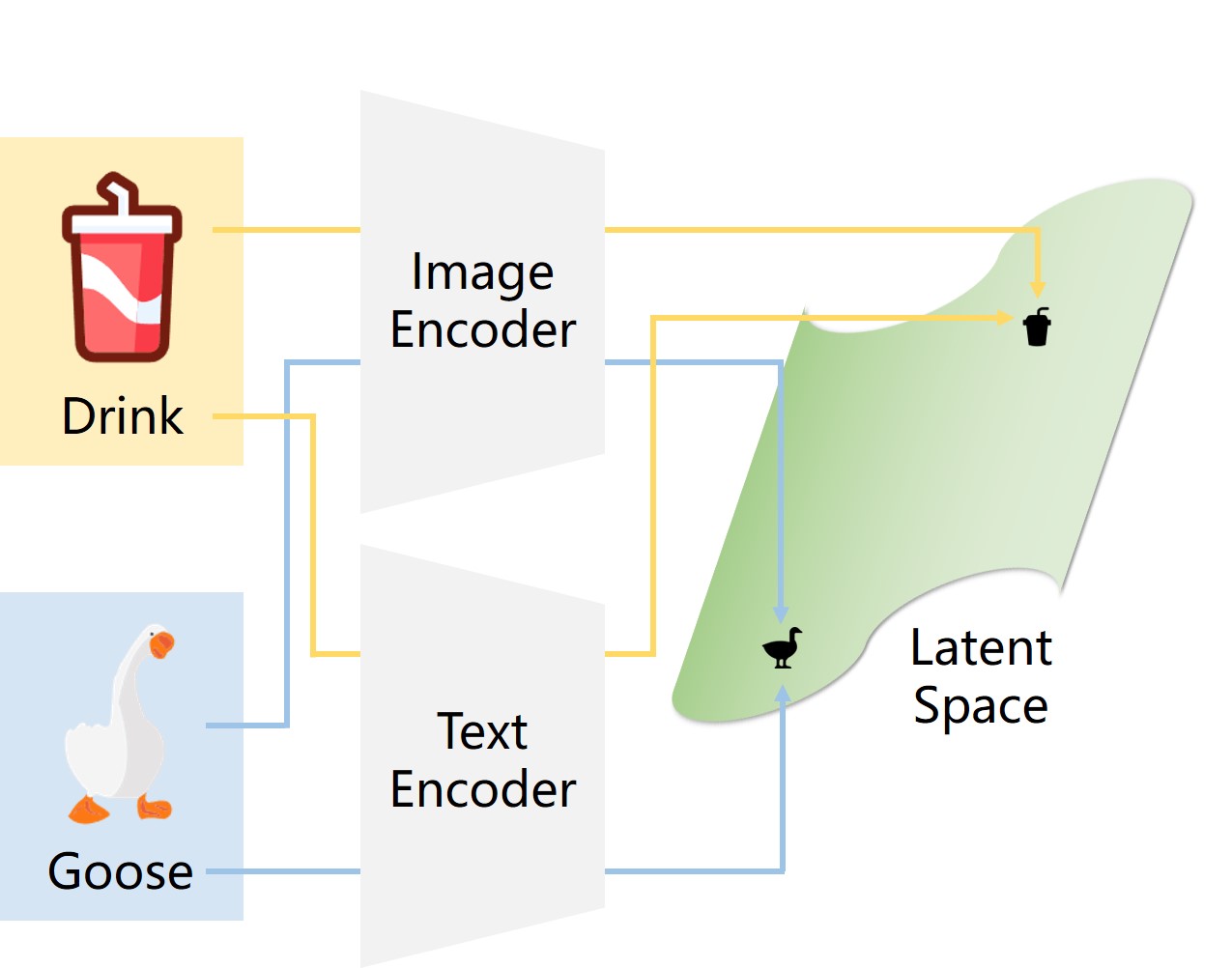}
            \caption{{The structure of dual encoder.}}
            \label{fig4}
        \end{wrapfigure}
        each modality independently. The core idea of this architecture is to guide the learning process of the two encoders in synchronization through semantic similarity metrics by leveraging contrastive learning methods. Then, the output feature vectors of different encoders can be projected into the same vector space. Specifically, this model is based on the hypothesis that if the feature vectors output by the two encoders belong to the same feature space, the feature vectors with paired labels should be closer in the vector space and vice versa. Through this alignment method, we can expect that the output results of encoders for different modalities describing similar objects or scenes will be close enough. And even in an ideal state, they will converge to the same point in the feature space. The key to achieving this goal is to construct a reasonable model architecture and thoroughly train it on a large-scale dataset. Figure~\ref{fig4} shows the processing logic of the dual encoder.
    \end{minipage}
    
    % \begin{figure}[!t]
    %     \centering
    %     \includegraphics[width=0.5\linewidth]{./img/4.jpg}
    %     \caption{The structure of dual encoder.}
    %     \label{fig4}
    % \end{figure}

    % \begin{minipage}{0.94\textwidth}
    %     \begin{wrapfigure}{r}{0.4\textwidth}
    %         \centering
    %         \includegraphics[width=0.4\textwidth]{./img/4.jpg}
    %         \caption{{The structure of dual encoder.}}
    %         \label{fig4}
    %     \end{wrapfigure}
    %     {However, the cost of one-to-one aligning each pair of modalities would be very high. Besides, obtaining a dataset with each pair of modalities aligned is also challenging. To this end, some researchers have proposed the bridging alignment (also known as binding alignment) strategy. This method matches all other modalities with a central modality, thereby achieving the alignment of all modalities in the semantic space. For example, ImageBind~\cite{girdhar2023imagebind} takes the image as the central modality, while CoDi~\cite{tang2024any} takes text as the central modality. In this way, they effectively simplify the training process of multi-modality alignment and improve the practicality and efficiency of the model. In contrastive learning, the class imbalance in long-tailed datasets can lead to reduced effectiveness, as most contrastive pairs consist of major classes instead of tail classes, making the model ineffective for the minor classes. Therefore,~\cite{du2024probabilistic} proposed a method to solve this problem.}
    % \end{minipage}
    However, the cost of one-to-one aligning each pair of modalities would be very high. Besides, obtaining a dataset with each pair of modalities aligned is also challenging. To this end, some researchers have proposed the bridging alignment (also known as binding alignment) strategy. This method matches all other modalities with a central modality, thereby achieving the alignment of all modalities in the semantic space. For example, ImageBind~\cite{girdhar2023imagebind} takes the image as the central modality, while CoDi~\cite{tang2024any} takes text as the central modality. In this way, they effectively simplify the training process of multi-modality alignment and improve the practicality and efficiency of the model. In contrastive learning, the class imbalance in long-tailed datasets can lead to reduced effectiveness, as most contrastive pairs consist of major classes instead of tail classes, making the model ineffective for the minor classes. Therefore,~\cite{du2024probabilistic} proposed a method to solve this problem.

\end{enumerate}

\subsection{{Gain of scaling up}}
% 随着AI模型的的训练规模逐渐增大，其在各种任务中的能力逐渐展现出了惊人的飞跃。当AI模型的大小上升到了FM级别时，研究者们引出了scaling law和emergent phenomenon来展示训练规模跟模型性能之间的量变与质变关系。
{
As the training scale of artificial intelligence (AI) models continues to grow, their capabilities in various tasks demonstrated significant leaps in performance. When the model size reaches the FM level, researchers have introduced the concepts of scaling laws and emergent phenomena to illustrate the quantitative and qualitative relationships between training scale and model performance.
}

% Scaling law
{
Scaling law refers to the mathematical pattern of how the performance of a system changes as the scale of the system increases. In AI, especially in the research and application of FMs, scaling law describes rules and phenomena about how model performance changes as the model scale expands, including the number of parameters, dataset size, and computational resources. It uses quantitative analysis methods to reveal the intrinsic mechanism of the performance improvement of FMs.
}

{
\cite{tay2022scaling} discussed how the inductive biases of different models affect the relationship between model scale expansion and performance. They found that model architecture is indeed one of the key factors affecting the benefits of model expansion. They also pointed out that although the standard transformer architecture may not always achieve the best performance, it does exhibit the best scalability. In computer vision~\cite{zhai2022scaling} and natural language process~\cite{kaplan2020scaling}, models based on the transformer architecture have shown an exponential relationship between model scale and model performance.
}

{
Besides,~\cite{chung2024scaling} examined the impact of the number of downstream tasks and model scale on the performance of instruction fine-tuning. They fine-tuned the models on various tasks by a multi-task joint training method. As a result, the language model could learn a broader language representation and knowledge, enhancing its generalization ability on unseen tasks. During the joint training process, knowledge transfer between different tasks is promoted through parameter sharing, significantly improving the model's generalization ability and performance. In addition, joint training also reduces the time and computational resources required to train each task individually, improving training efficiency. This phenomenon of model performance improving as task diversity increases is a manifestation of the scaling law.~\cite{iyer2022opt} verified the phenomenon of model performance increasing with the number of tasks on the large-scale benchmark OPT-IML Bench. Additionally, some studies have separately provided the model performance of natural language models~\cite{kaplan2020scaling} and autoregressive generative models of various modalities~\cite{henighan2020scaling} at different scales.
}

{
Although there is no unified form for the quantitative representation of the scaling law, it can be generally represented as an exponential relationship between the model loss function and the model's parameters, dataset size, and computational resources. We use the loss function $L(\cdot)$ to characterize the model's performance, where a smaller loss function value represents better model performance. Eq.~(\ref{scalling_1}) describes the performance of a model with a given number of parameters trained to convergence on a sufficiently large dataset, {where $L(N)$ is the loss function, $N$ is the number of trainable parameters of the model, $N_c$ is a constant, and $\alpha_N$ is the power law exponent.} Eq.~(\ref{scalling_2}) provides the performance of a suitably sized model trained on a sufficiently large dataset under a given computational resource constraint, where $L(C)$ is the loss function, $C$ is the given computational resource, $C_c$ is a constant, and $\alpha_C$ is the power law exponent. Eq.~(\ref{scalling_3}) describes the performance of an FM trained with an early stopping strategy on a given dataset size, where $L(D)$ is the loss function, $D$ is the dataset size (in tokens), $D_c$ is a constant, and $\alpha_D$ is the power law exponent.
}

{
\begin{align}
    L(N) &\propto \left(\frac{N_c}{N}\right)^{\alpha_N} \label{scalling_1} , \\
    L(C) &\propto \left(\frac{C_c}{C}\right)^{\alpha_C} \label{scalling_2} , \\
    L(D) &\propto \left(\frac{D_c}{D}\right)^{\alpha_D} \label{scalling_3} .
\end{align}
}

{
According to the above equations, when not constrained by other conditions, the model's loss function decreases exponentially as the number of parameters, computational resources, and training data volume increase. This means that by increasing the model's parameters, investing more in computational resources, and expanding the training data volume, the model's performance can also be exponentially improved.
}

% Emergent phenomenon
{
The scaling law reveals that the scaling up of model size can lead to incremental improvements in model performance. On the contrary, the emergent phenomenon refers to the new properties exhibited by the model after reaching a critical point of scale expansion, one of which is a significant improvement in model performance~\cite{wei2022emergent}.
}

{
The emergent phenomenon essentially reveals the source of the superior performance of FMs. In deep learning, especially in the domain of LLM, the emergent phenomenon has been widely observed. For example, models like LLaMA have demonstrated exceptional comprehension, generation capabilities, and even a certain degree of logical reasoning ability, which small language models cannot achieve. As the model scale increases, it can have more parameters and a more complex structure, allowing it to capture the complex features and patterns in the data. FMs often exhibit strong generalization capabilities because their abundant parameters can store rich knowledge, enabling them to make accurate inferences and predictions on unseen data. It can provide them adaptability and universality to different tasks and even allows the model to truly learn the underlying principles and reasoning methods in the data. \cite{wei2022emergent} suggested that the task and the prompting method used can influence the emergence point of the emergent phenomenon in LLMs. Specifically, using a chain-of-thought prompting approach can significantly enhance the LLMs' ability to handle complex reasoning tasks~\cite{wei2022chain}, thereby bringing forward the emergence point of the emergent phenomenon.
}

{
When building domain-specific FMs, developers must select an appropriate model scale, considering both the requirements of potential downstream tasks and user prompting patterns. An increase in the number of model parameters raises the demand for computational resources and the risk of overfitting. Consequently, there is a limit to the extent that parameters can be increased. The trade-offs highlighted by this phenomenon are crucial for model deployment, offering significant insights for both model design and application.
}

\subsection{{Performance comparison}}
{
In this section, we present the evaluation results of current multi-modality FMs to assist readers in selecting the appropriate model for their needs. We summarize some common vision-language tasks in Table \ref{tab:result} according to ~\cite{wu2023next,chen2025sharegpt4v,lin2024moe, chen2024internvl}. We select three mainstream vision-language tasks, including: 
\begin{itemize}
    \item \textbf{Image Captioning}: This task involves generating descriptive captions for images, demonstrating models' ability to understand visual content.
    \item \textbf{Image Question Answering}: In this task, models are required to answer questions about the content of images, demonstrating their ability to comprehend and reason about visual information.
    \item \textbf{Benchmark Toolkit}: This refers to a collection of standardized metrics and datasets used to evaluate the performance of multi-modal models across various tasks like conversation, complex reasoning, and detailed description, ensuring a consistent assessment framework.
\end{itemize}
}

{
In summary, InternVL-Chat~\cite{chen2024internvl} has the best results in Image Captioning and Image Question Answering tasks. While considering language tasks such as Benchmark Toolkits, the best results are provided by NExT-GPT~\cite{wu2023next}.
}

\begin{table*}[!t]
\footnotesize
% \small
\caption{{Performance of various large multi-modality models on vision-language tasks}}
% . The best results are \textbf{bold}, and the second best results are \underline{underlined}

\setlength{\tabcolsep}{1pt} % 列间距，默认是6pt
\fontsize{9pt}{12pt}\selectfont % 字号，行距

\label{tab:result}
\def\tabblank{\hspace*{10mm}}
\begin{adjustbox}{width=\textwidth}

% \color{blue}

\begin{tabular}{lc|ccc|cccccc|ccccccc}
\toprule
\multirow{2}*{\textbf{Model}} & \multirow{2}*{\textbf{LLM}} & \multicolumn{3}{c|}{\textbf{Image Captioning}} & \multicolumn{6}{c|}{\textbf{Image Question Answering}} & \multicolumn{7}{c}{\textbf{Benchmark Toolkit}} \\
& & \textbf{NoCaps~\cite{agrawal2019nocaps}} & \textbf{Flickr30K~\cite{young2014image}} & \textbf{COCO~\cite{karpathy2015deep}} & 
\textbf{VQA$^{v2}$~\cite{goyal2017making}} & \textbf{GQA~\cite{hudson2019gqa}} & \textbf{VizWiz~\cite{gurari2018vizwiz}} & \textbf{SQA$^I$~\cite{lu2022learn}} & \textbf{VQA$^{T}$~\cite{singh2019towards}} & \textbf{OKVQA~\cite{marino2019ok}} & 
\textbf{POPE~\cite{li2023evaluating}} & \textbf{MME~\cite{yin2023survey}} & \textbf{MMB~\cite{liu2025mmbench}} & \textbf{LLAVA$^W$~\cite{liu2024visual}} & \textbf{SEED~\cite{li2023seed}} & \textbf{MM-Vet~\cite{yu2023mm}} & \textbf{QBench~\cite{wu2023q}} \\
\midrule

mPLUG-Owl ~\cite{ye2023mplug} & LLaMA-7B ~\cite{touvron2023llama}    & 117.0     & 80.3     & 119.3 & -     & -    & 39.0 & -    & - & -      & -     & -     & 46.6 & - & 34.0 & - & -\\
Emu  ~\cite{sun2023generative}     & LLaMA-7B ~\cite{touvron2023llama}     & -     & -     & 117.7 & 40.0  & -    & 35.4 & -    & - & 34.7  & -     & -     & - & -     & -     & - & -\\
I-80B ~\cite{laurenccon2024obelics} & LLaMA-65B  ~\cite{touvron2023llama}        & 65.0     & 53.7     & 91.8     & 60.0  & 45.2 & 36.0 & -    & 30.9 & -    & -     & -      & 54.5  & -     & -     & - & -\\
% IDEFICS-80B & LLaMA-65B ~\cite{touvron2023llama}  & -     & -     & -     & 60.0  & -    & 36.0 & -    & -    & -      & -  & -  & 54.5 & -  & -  & -  & - \\
LLAVA ~\cite{liu2024visual}  & LLaMA2-7B  ~\cite{touvron2023llama2}     & 120.7     & 82.7     & -     & -  & -    & -     & -    & -    & -  &- & - & 36.2 & -  &- & - & -\\

MobileVLM ~\cite{chu2023mobilevlm} & MobileLLaMa-2.7B  ~\cite{kan2024mobile}    & -     & -     & -     & -     & 59.0 & -    & 61.0 & 47.5 & -    & 84.9  & 1288.9 & 59.6  & -     & -     & - & -\\

DREAMLLM ~\cite{dong2023dreamllm} & Vicuna-7B ~\cite{chiang2023vicuna}     & -     & -     & 115.4 & 56.6  & -    & 45.8 & -    & - & \underline{44.3}   & -     & -  & 49.9 & -     & -     & -& -\\
Video-LLAVA ~\cite{lin2023video} & Vicuna-7B ~\cite{chiang2023vicuna}  & -     & -     & -     & 74.7  & -    & 48.1 & -    & -    & -      & -     & -  & 60.9  & -     & -     & -& -\\
NExT-GPT ~\cite{wu2023next}   & Vicuna-7B ~\cite{chiang2023vicuna}   & \underline{123.7} & 84.5  & \underline{124.9} & 66.7  & -    & 48.4 & -    & - & \textbf{52.1} & - & - & 58.0   & -    & 57.5  & - & -\\
ShareGPT4V-7B ~\cite{chen2025sharegpt4v} & Vicuna-7B ~\cite{chiang2023vicuna}  & -     & -     & -   & \underline{80.6} & - & \underline{57.2} & \underline{68.4} & - & - & - & \underline{1567.4} & \textbf{68.8} & \textbf{72.6} & \textbf{69.7} & \textbf{37.6} & \textbf{63.4} \\
LLAVA-1.5 ~\cite{liu2024improved} & Vicuna-7B ~\cite{chiang2023vicuna}     & -     & -     & -     & 78.5  & 62.0 & 50.0 & - & 58.2 & -    & \underline{85.9}  & 1510.7 & -  & -  & -     & - & - \\

LLAVA-1.5 ~\cite{liu2024improved} & Vicuna-13B ~\cite{chiang2023vicuna}     & -     & -     & -     & 80.0  & \underline{63.3} & 53.6 & \textbf{71.6} & 61.3 & -    & \underline{85.9}  & 1531.3 & \underline{67.7}  & \underline{70.7}  & \underline{68.2}     & \underline{35.4} & \underline{62.1} \\

InstructBLIP ~\cite{dai2023instructblip} & Vicuna-7B ~\cite{chiang2023vicuna}  & 123.1 & 82.4  & 102.2 & -     & -    & 34.5 & 60.5    & 50.1    & 33.9 & -      & -     & 36.0  & 60.9     & 53.4     & 26.2 & 56.7 \\

InstructBLIP ~\cite{dai2023instructblip} & Vicuna-13B ~\cite{chiang2023vicuna}  & 121.9 & 82.8  & - & -     & 49.5    & 33.4 & -    & 50.7    & - & 78.9      & 1212.8     & -  & -     & -     & - & - \\

Shikra  ~\cite{chen2023shikra} & Vicuna-13B  ~\cite{chiang2023vicuna}  & -     & 73.9     & 117.5     & 77.4  & -    & -     & -    & -    & -      & -  & -     & 58.8 & - & -& - & 54.7 \\

InternVL-Chat* ~\cite{chen2024internvl}    & Vicuna-13B  ~\cite{chiang2023vicuna}   & \textbf{126.2}     & \textbf{92.2}     & \textbf{146.2}     & \textbf{81.2}  & \textbf{66.6}    & \textbf{58.5} & -     & \underline{61.5}    & - & \textbf{87.6}  & \textbf{1586.4}  & - & - & - & - & - \\

Qwen-VL ~\cite{bai2023qwen_tr} & Qwen-7B  ~\cite{bai2023qwen_tr}    & 121.4     & \underline{85.8}     & -     & 78.8  & 59.3 & 35.2 & 67.1 & \textbf{63.8} & -    & -     & -      & 38.2  & -     &  56.3     & - & 59.4 \\
Qwen-VL-Chat ~\cite{bai2023qwen} & Qwen-7B  ~\cite{bai2023qwen_tr}  & 120.2     & 81.0     & -     & 78.2 & 57.5 & 38.9 & 68.2 & \underline{61.5} & - & - & 1487.5 & 60.6 & - & 58.2 & - & -\\

TinyGPT-V ~\cite{yuan2023tinygpt} & Phi2-2.7B ~\cite{javaheripi2023phi}    & -     & -     & -     & -     & 33.6 & 33.4 & -    & -    & -    & -     & -      & -     & -     & -     & - & - \\
LLAVA-Phi ~\cite{zhu2024llava} & Phi2-2.7B ~\cite{javaheripi2023phi} & -     & -     & -     & 71.4  & -    & 35.9 & \underline{68.4} & 48.6 & -    & 85.0  & 1335.1 & 59.8  & -     & -     & 28.9 & -\\

BLIP-2 ~\cite{li2022blip}    & FLAN-T5  ~\cite{chung2024scaling}   & 103.9     & 71.6     & -     & 41.0  & 41.0    & 19.6 & 61.0    & 42.5    & - & 85.3  & 1293.8  & - & 38.1 & 46.4 & 22.4 & - \\

\bottomrule
\end{tabular}
\end{adjustbox}

\footnotesize{{*: InternVL-Chat with IViT-6B as the visual encoder, QLLaMA as the glue layer and 13B trained parameters}}

\footnotesize{{\textbf{Bold}: The best results}}

\footnotesize{{\underline{Underlined}: The second best results}}

\end{table*}

\section{Key technologies for building domain-specific foundation models}
This section delves into the technical pathways toward customizing domain-specific FMs. We will explain how to flexibly select and combine the appropriate modules from five key components --- MEs, IPs, BCs, OPs, and MDs based on the specific requirements of different domains. Additionally, we will analyze concrete cases to help readers better understand and apply the methodologies discussed in this section.

We can categorize the customization of domain-specific FMs into three levels, ranging from low to high level of customization (i.e., from high to low reliance on general-purpose FMs or pre-trained modules): 

\begin{enumerate}
    \item Domain-specific enhancement based on general-purpose FMs.
    \item Customization of the FM based on pre-trained modules.
    \item Construction of the FM without pre-trained modules.
\end{enumerate}

Table~\ref{tab1} summarizes the characteristics of these three domain-specific FM customization methods.

\begin{table*}[t!]
    \footnotesize
    \caption{Customization methods of domain-specific FMs}
    \label{tab1}
    \def\tabblank{\hspace*{10mm}}
    
    \begin{tabularx}{\textwidth}{@{\extracolsep{\fill}}ccccc}
    \toprule
    Customization Method & \makecell[c]{Customization Degree} & \makecell[c]{Difficulty} & Flexibility & \makecell[c]{Computation} \\
    \hline
    \makecell[c]{Based on general-purpose\\FMs} & \makecell[c]{Low, only the input method of\\domain knowledge is customized} & Low & Low & Low \\
    \hline
    \makecell[c]{Based on pre-trained modules} & \makecell[c]{Medium, some modules can be customized} & Medium & Medium & Medium \\
    \hline
    \makecell[c]{Without pre-trained modules} & \makecell[c]{High, each module can be customized} & High & High & High \\
    \bottomrule
    \end{tabularx}
\end{table*}

\subsection{Domain-specific enhancement based on general-purpose foundation model}
General-purpose FMs offer comprehensive capabilities, making them suitable for various task scenarios. When such a general-purpose FM can fully handle the required data modalities, any modifications to its underlying architecture for model developers are rendered unnecessary. Instead, they can focus on implementing domain-specific enhancements.

Depending on whether the domain-specific enhancement requires altering the parameters of the FM, we can further divide this into plug-and-play domain-specific enhancement and fine-tuning-based domain-specific enhancement. Table~\ref{tab2} categorizes and summarizes the methods of domain-specific enhancement based on a full-architecture general-purpose FM.

\begin{table*}[t!]
    \footnotesize
    \caption{Specific-domain enhancement with the entire general-purpose FMs}
    \label{tab2}
    \tabcolsep 11.5pt %space between two columns.
    
    \begin{tabular}{c|c|c|c|c|c|c}
    \toprule
    \multicolumn{3}{c|}{Customization Method} & \makecell[c]{Parameter \\ Modification} & \makecell[c]{Addition \\ of New \\ Modules} & \makecell[c]{Domain \\ Knowledge \\ Provider} & Specific Techniques \\
    \hline
    \multirow{7}*{\makecell[c]{Plug\\-and\\-play}} & \multirow{3}*{\makecell[c]{Utilizing \\ Existing \\ Knowledge}} & \makecell[c]{Hard \\ Prompting} & No & No & Deployer & PET~\cite{schick2020exploiting} \\
    \cline{3-7}
    & & \makecell[c]{Soft \\ Prompting} & No & Yes & Deployer & \makecell[c]{Prefix Tuning~\cite{li2021prefix}, \\P-tuning~\cite{liu2023gpt}} \\
    \cline{2-7}
    & \multirow{4}*{\makecell[c]{Adding New \\ Knowledge}} & {Prompts} & {No} & {No} & {User} & \makecell[c]{LongRoPE~\cite{ding2024longrope}, \\Transformer-XL~\cite{dai2019transformer}} \\
    \cline{3-7}
    & & {\makecell[c]{External \\ Knowledge \\ Base}} & {No} & {Yes} & {Deployer} & \makecell[c]{RAG~\cite{lewis2020retrieval}, \\GraphRAG~\cite{edge2024local}, \\RankRAG~\cite{yu2024rankrag}}\\
    \hline
    \multirow{5}*{\makecell[c]{\makecell[c]{Fine\\-tuning\\-based}}} &  \multicolumn{2}{c|}{Adapter-Based} & {Yes} & {Yes} & {Deployer} & \makecell[c]{Adapter~\cite{houlsby2019parameter}, \\AdapterFusion~\cite{pfeiffer2020adapterfusion}, \\IA3~\cite{liu2022few}, Model \\Reprogramming~\cite{chen2024model}} \\
    \cline{2-7}
    & \multicolumn{2}{c|}{\makecell[c]{Low-Rank Matrix \\ Decomposition}} & {Yes} &{Yes} & {Deployer} & \makecell[c]{LoRA~\cite{hu2021lora}, \\LoHa~\cite{kim2016hadamard}, \\LoKr~\cite{hackbusch2006low}} \\
    \cline{2-7}
    & \multicolumn{2}{c|}{\makecell[c]{Full Parameter \\ Fine-tuning}} & Yes & No & Deployer & PEFT~\cite{ding2023parameter} \\
    \bottomrule
    \end{tabular}
\end{table*}

\subsubsection{Plug-and-play domain-specific enhancement}
\begin{figure}[t!]
    \centering
    \begin{minipage}[c]{0.32\textwidth}
        \centering
        \includegraphics[height=3.1cm]{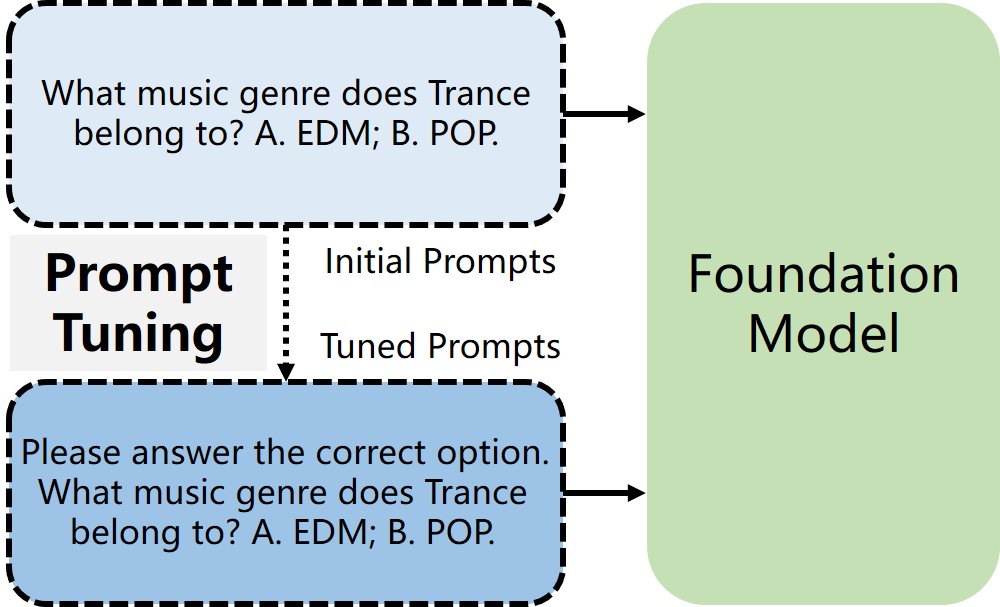}
        
        (a)
    \end{minipage}
    \hfill
    \begin{minipage}[c]{0.32\textwidth}
        \centering
        \includegraphics[height=3.1cm]{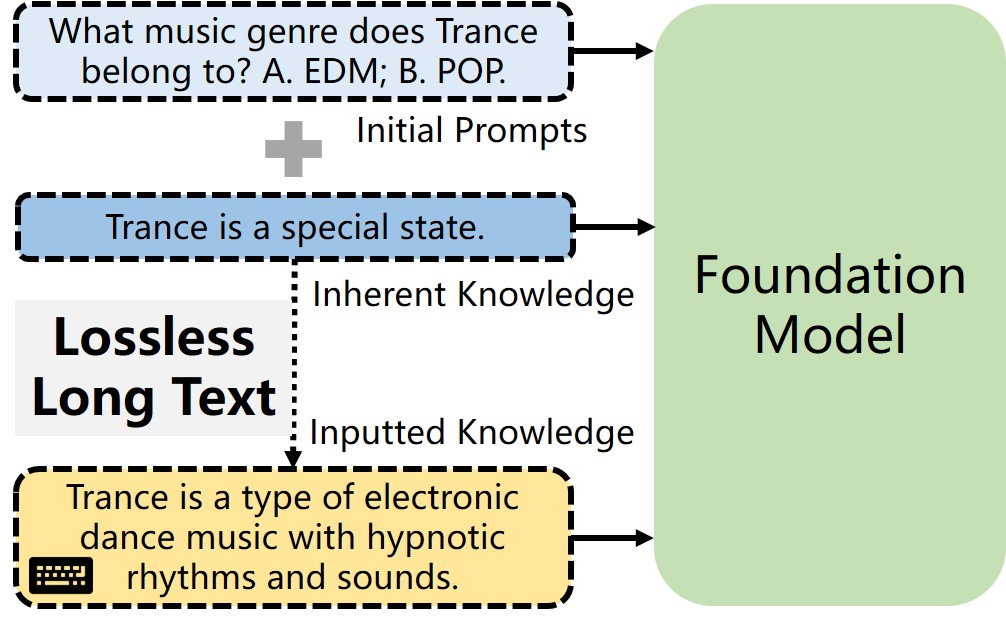}
        
        (b)
    \end{minipage}
    \hfill
    \begin{minipage}[c]{0.32\textwidth}
        \centering
        \includegraphics[height=3.1cm]{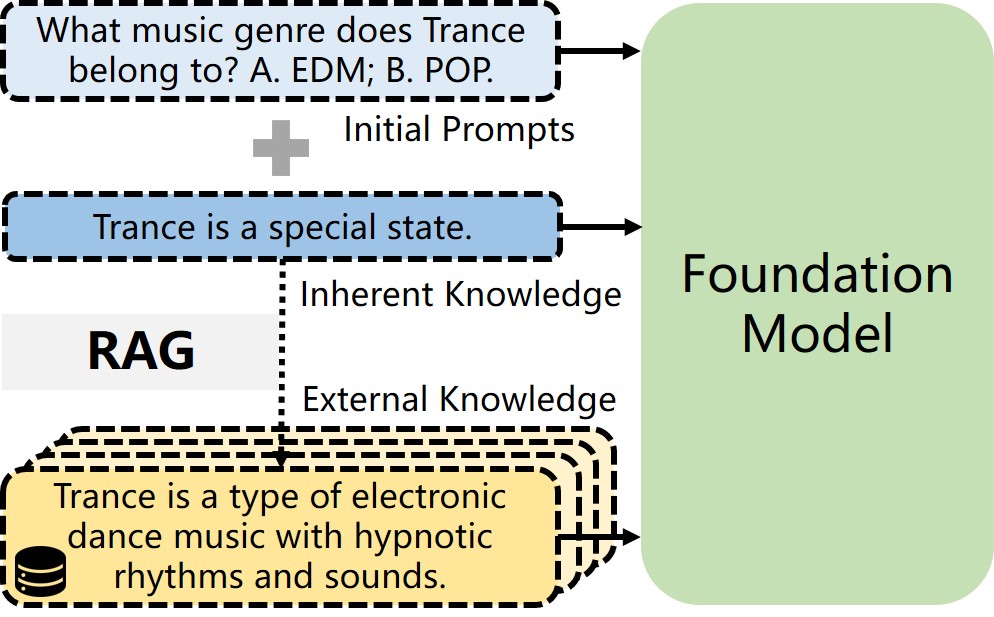}
        
        (c)
    \end{minipage}
    
    \caption{Plug-and-play domain-specific enhancement. (a) Invoking existing knowledge; (b) Knowledge embedding by prompts; (c) Knowledge embedding by external knowledge base.}
    \label{fig5}
\end{figure}

The generality, generalization capabilities, and reasoning abilities of general-purpose FMs enable them to serve as the foundation for domain-specific models. To achieve plug-and-play domain enhancement without modifying the parameters of the FM, two approaches can be utilized: \textbf{leveraging existing knowledge} or \textbf{embedding new knowledge}. The first approach aims to leverage domain knowledge already stored within the general-purpose FM, as illustrated in Figure~\ref{fig5} (a). The second approach, embedding new knowledge, involves equipping the FM with the ability to handle domain tasks by introducing domain-specific knowledge. This can be divided into two methods: embedding knowledge through prompts and embedding knowledge via an external knowledge base. These methods are depicted in Figure~\ref{fig5} (b) and (c) respectively. The following sections will provide a detailed explanation of these techniques.

\begin{enumerate}
    \item \textbf{Invoking existing knowledge for domain enhancement}: A general-purpose FM may have already enclosed domain knowledge during the training process. Prompt tuning improves prompts to better invoke the model's inherent domain knowledge, where "tuning" refers to the optimization of prompts. Specifically, it involves inserting a carefully crafted prompt into the input data as context to improve the generated output. These carefully crafted prompts can be natural language descriptions, examples, rules, or other text or embedding vectors that guide the model to understand the task requirements. The model will consider these carefully crafted prompts to produce task-relevant results when generating outputs. Prompt tuning can be categorized into hard prompts and soft prompts:
    \begin{enumerate}
        \item \textbf{Hard prompts}: Hard prompt methods are common techniques in natural language processing (NLP). They use interpretable and reusable handcrafted words and tokens to guide the language model's output. Hard prompts are typically manually designed and tailored for specific tasks, making them difficult to modify. PET (Pattern Exploiting Training)~\cite{schick2020exploiting} is a classic hard prompt learning method that models questions as cloze tasks and optimizes the final output words. This method trains the model on a small amount of supervised data and performs ensembling predictions on unsupervised data to guide the model.
        {
        \item \textbf{Soft prompts}: Designing hard prompts requires experimental exploration and expertise, and manually designed prompts may not align well with the model's data processing methods. To simplify this process and enhance the flexibility of prompt tuning, researchers proposed soft prompt-based tuning methods. Prefix Tuning~\cite{li2021prefix} is a form of soft prompt tuning that adapts to specific downstream tasks by adding learnable prefix vectors (soft prompts) to the beginning of the input sequence. As part of the input, these prefix vectors guide the model's output to meet task requirements. The advantage of prefix tuning is that it only updates these prefix vectors rather than the model's parameters, significantly reducing computational and storage resource demands while retaining the rich knowledge learned by the pre-trained model. Building on prefix tuning, researchers introduced the P-tuning method~\cite{liu2023gpt}. P-tuning replaces fixed or manually designed words and tokens with learnable soft prompts. Its core idea is to treat prompts as part of the model that can be learned, allowing the model to learn not only how to respond to given tasks but also how to generate the best prompts. These soft prompts are typically a series of embedding vectors processed alongside the actual text input. Through end-to-end training, the model automatically learns to adjust these embedding vectors for better task performance. P-tuning combines the parameter efficiency of prefix tuning with the flexibility of traditional hard prompt tuning. Soft prompts give the model greater freedom to generate answers, potentially producing more diverse outputs and increasing the risk of generating inaccurate or irrelevant responses.
        }
    \end{enumerate}
    
    \item \textbf{Knowledge embedding for domain enhancement}: When the existing knowledge of a general-purpose FM is insufficient to solve domain tasks, introducing new knowledge by embedding additional background information can achieve higher-quality output. This method is known as knowledge embedding for domain enhancement.
    \begin{enumerate}
        \item \textbf{Knowledge embedding by prompts}: Prompts, serving as a direct interface between the user and the large language model, can be used to incorporate domain knowledge. However, the method of knowledge embedding through prompts has a significant limitation: the amount of embedded domain knowledge is restricted by the maximum prompt length of the model. The limitation on the model's ability to process longer text inputs stems from three core issues of the Transformer architecture:
        \begin{itemize}
           \item \textbf{Limitations of positional encoding}: Transformer models typically generate fixed-length positional encodings using sine and cosine functions, where each position in the sequence is uniquely encoded. However, when the sequence length exceeds the maximum length used during training, the model cannot effectively handle the additional text because it cannot generate valid encodings for the new positions.
            \item \textbf{Resource consumption of the attention mechanism}: The attention mechanism is the core of Transformer models, allowing the model to compute attention weights for each element in the sequence. However, as the sequence length increases, this mechanism's computational complexity and memory requirements grow quadratically, leading to significant resource consumption.
            \item \textbf{Long-distance dependency issue}: When handling long sequences, the Transformer needs to span a large number of input tokens, which often results in problems such as gradient vanishing or exploding. This makes it difficult for the model to capture dependencies between elements far apart in the sequence.
        \end{itemize}
        To address the above issues, Lossless Long Text technology has emerged. It aims to enhance the model's ability to process long texts that exceed its input length limitations, allowing users to input a large amount of domain knowledge directly through prompts into the large language model as contextual information for domain enhancement. Lossless Long Text technology expands the long-text input capability of large language models in two directions: extrapolation and interpolation:
        \begin{enumerate}
            \item \textbf{Extrapolation}: Extrapolation involves extending the model's context window to handle new texts that exceed the length of the training data. This typically involves improving the positional encoding mechanism so the model can understand and process longer sequences. Longformer~\cite{beltagy2020longformer} extends the ability to handle long texts effectively by combining local and global attention mechanisms; BigBird~\cite{zaheer2020big} uses sparse attention mechanisms and reversible layers to extrapolate the model's long-sequence processing capabilities; LongRoPE~\cite{ding2024longrope} improves positional encoding by introducing rotational transformations in self-attention, allowing the model to handle long-distance dependencies and support inputs up to two million tokens without impacting computational efficiency.
            \item \textbf{Interpolation}: Interpolation refers to enhancing the model's ability to process long texts within its existing sequence length capacity by adjusting and optimizing the attention mechanism. This typically involves improvements to the attention mechanism so the model can more effectively handle long-distance information. The BERT model~\cite{devlin2018bert} enhances text understanding through pre-training with a bidirectional Transformer. XLNet~\cite{yang2019xlnet} improves long text processing by using permutation language modeling and generalized autoregressive pre-training to enhance the model's internal representations. Transformer-XL~\cite{dai2019transformer} is an improved Transformer model that introduces a recurrence mechanism to address the issue of gradient vanishing in long text processing. This allows the model to retain information from previous sequences while processing the current sequence, thereby better understanding and generating long text content.
        \end{enumerate}

        \begin{figure}[!t]
            \centering
            \centerline{\includegraphics[width=1.0\linewidth]{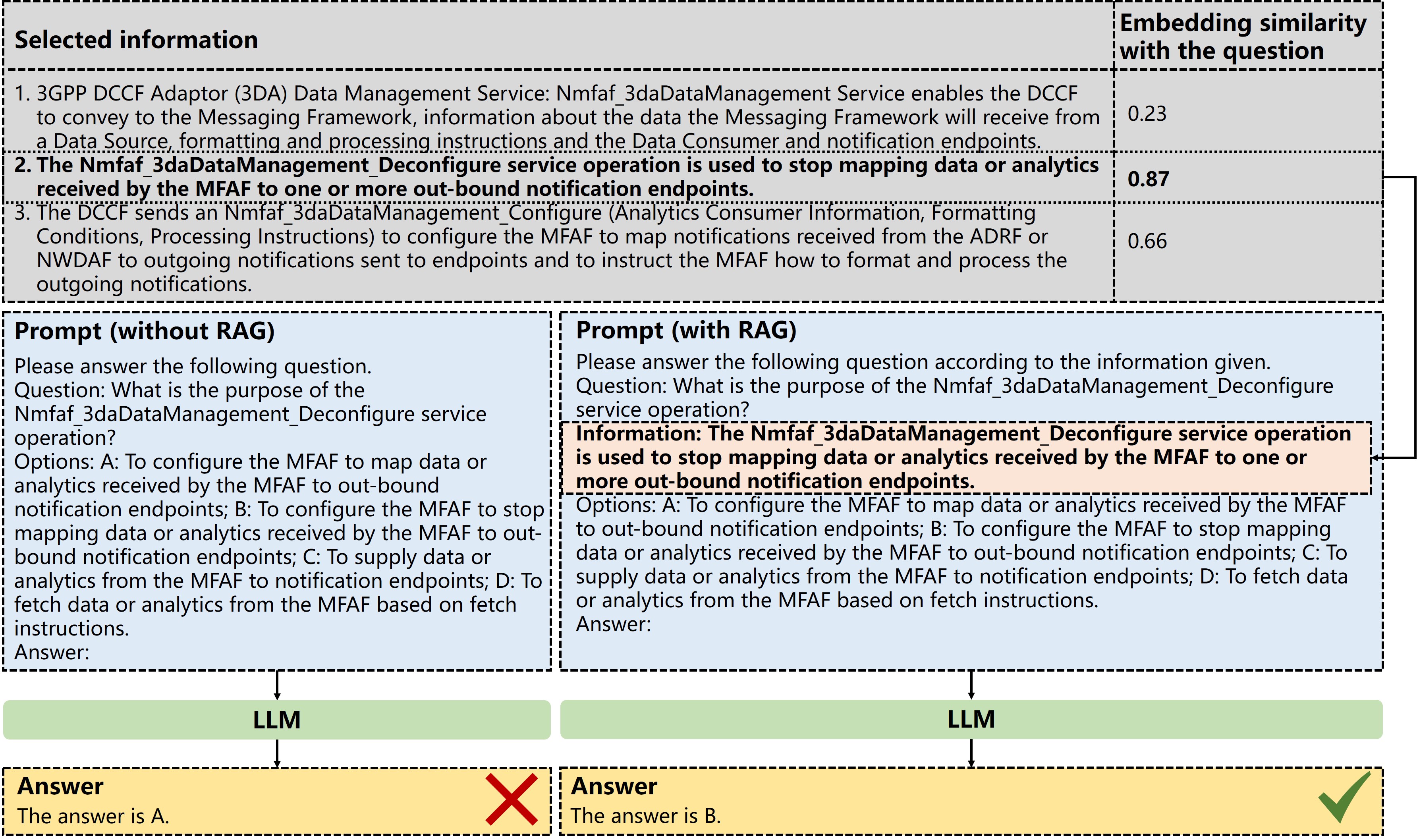}}
            \caption{An example of using LLM to answer multiple-choice communication questions. RAG improves the accuracy of answers by retrieving additional knowledge for the LLM from a knowledge base based on the relevance to the question text. The data used in the example is from \cite{maatouk2023teleqna}.}
            \label{RAG_example}
        \end{figure}
        
        \item \textbf{Knowledge embedding by external knowledge base}: In practical applications, users may be unable to provide sufficient domain knowledge to enhance a general-purpose FM. To address this issue, model deployers can augment the general-purpose FM with a dedicated domain knowledge base. This method allows the general-purpose FM to reference this external knowledge base when generating answers or performing tasks, thereby obtaining the necessary domain information and context to provide more accurate and targeted responses or solutions. {Retrieval-Augmented Generation (RAG)~\cite{lewis2020retrieval, ding2024survey, gao2023retrieval} technology was developed to achieve this purpose. RAG technology aims to enhance the language model's generation capabilities by leveraging an external document base without retraining the model. It is particularly suitable for tasks requiring a customizable dynamic knowledge base, such as question-answering, text summarization, and fact-checking. The core of RAG technology is the integration of a retrieval component that can quickly find information relevant to the current task in a large document database during the generation process. Once the relevant documents are retrieved, this information is used as additional contextual information to aid the generation process. The advantage of RAG technology is that it combines the generation capabilities of large language models with the knowledge provided by external retrieval systems without requiring domain knowledge themselves. Furthermore, because the external knowledge base can be replaced as needed, RAG technology offers high flexibility and adaptability. To have an intuitive understanding of RAG, we present an example in Figure~\ref{RAG_example}.}
        
        In RAG, the two core challenges are retrieving useful information from the database and organizing that information to augment the generation of FMs. For the retrieval task, the earliest and most naive approach uses sparse retrieval that directly matches data based on raw data like BM25~\cite{chen2017reading}. Inspired by the field of information retrieval, dense retrieval like DPR~\cite{karpukhin2020dense} has been proposed, which projects raw data into a high-dimensional space, potentially helping to capture semantic similarity better. However, a significant challenge is that even if the retrieved data is highly related to the original input in the semantic space, we cannot guarantee that it will help the model generation due to issues like polysemy. To address this problem, some researchers have introduced other technologies like knowledge graphs to aid the retrieval process~\cite{edge2024local}. \cite{cuconasu2024power} showed that adding noise can help enhance the model performance compared to traditional dense retrieval methods. Additionally, the organization of the retrieved information can directly influence the output quality. For example, providing a ranking of the retrieved data~\cite{yu2024rankrag} can improve model performance.
    \end{enumerate}
\end{enumerate}

While the aforementioned technologies were initially proposed to enhance large language models in specific domains, their applications are not limited to language models. With the development of FM domains, these technologies are expected to be extended to FMs of other modalities.

\subsubsection{Domain-specific enhancement based on fine-tuning}

When plug-and-play domain enhancement techniques are difficult to implement or require embedding too much domain knowledge into the general-purpose FM, or when deep modifications to the general-purpose FM are necessary, we can turn to the strategy of domain enhancement based on fine-tuning. This strategy aims to customize the required domain-specific FM while preserving the pre-trained knowledge of the general-purpose FM as much as possible by specific domain enhancement~\cite{han2024parameter}.

\begin{figure}[t!]
    \centering
    \begin{minipage}[c]{0.28\textwidth}
        \centering
        \includegraphics[height=6.5cm]{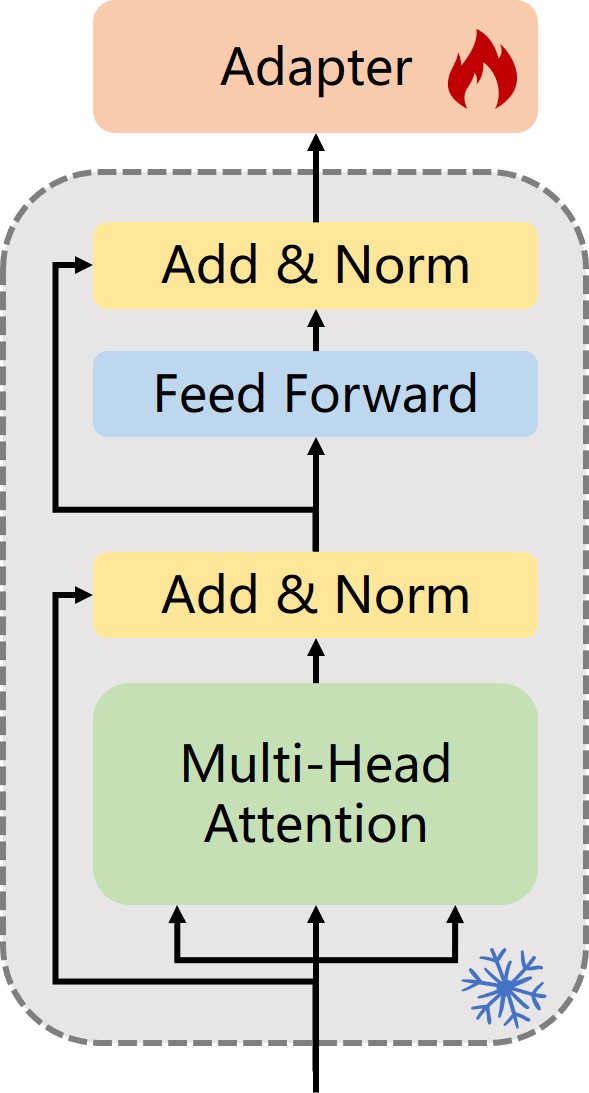}
        
        (a)
    \end{minipage}
    \hfill
    \begin{minipage}[c]{0.45\textwidth}
        \centering
        \includegraphics[height=6.5cm]{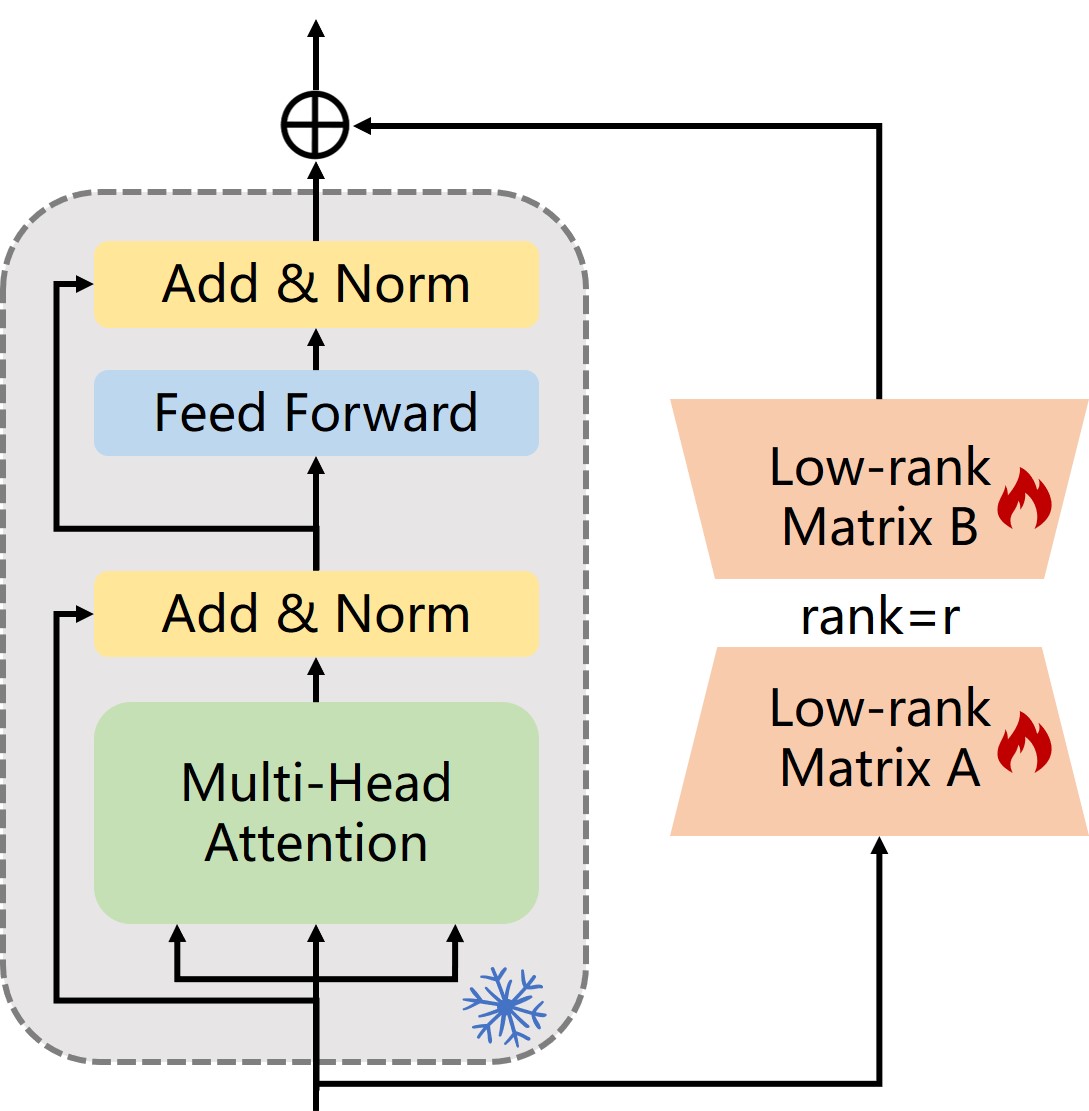}
        
        (b)
    \end{minipage}
    \hfill
    \begin{minipage}[c]{0.25\textwidth}
        \centering
        \includegraphics[height=6.5cm]{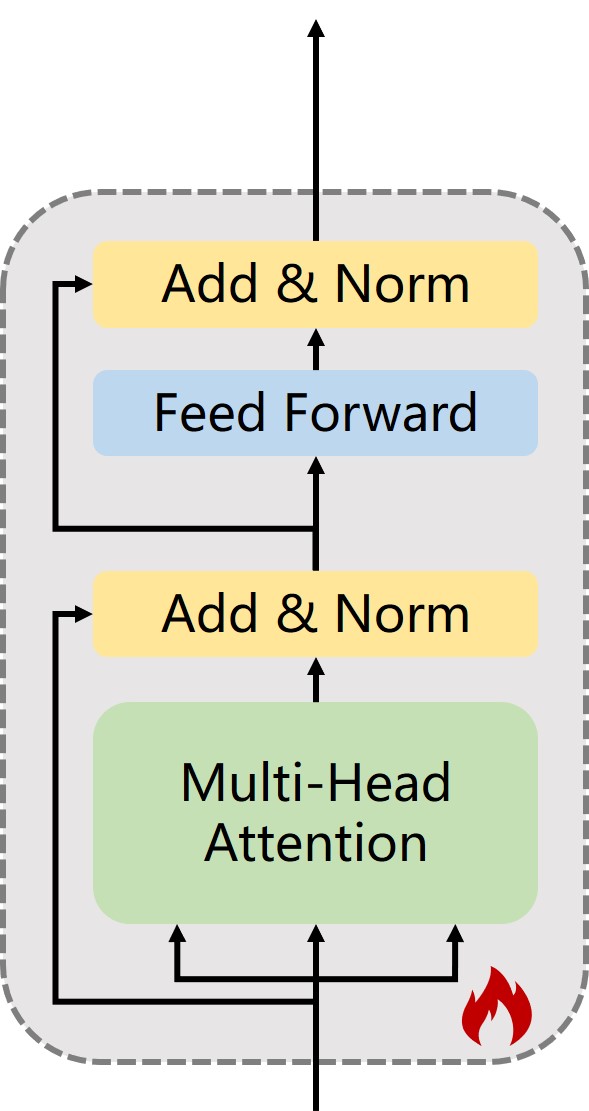}
        
        (c)
    \end{minipage}
    
    \caption{{Fine-tuning-based domain-specific enhancement. (a) Adapter-based tuning; (b) Low-rank-decomposition-based tuning; (c) Full fine-tuning.}}
    \label{fig6}
\end{figure}

Fine-tuning techniques can be divided into three main types: adapter-based fine-tuning, low-rank matrix decomposition-based fine-tuning, and full-parameter fine-tuning. Figure~\ref{fig6} (a), (b), and (c) illustrate these three technical pathways, respectively. In the following sections, this paper will elaborate on these techniques, sorted from low to high resource requirements and the complexity needed for fine-tuning.

\begin{enumerate}
    \item \textbf{Adapter-based fine-tuning}: Adapter-based fine-tuning~\cite{houlsby2019parameter} is a method that inserts small trainable adapter modules into pre-trained models, aiming to efficiently adapt the model to specific downstream tasks. During fine-tuning, only the parameters of adapter modules are updated, while the original parameters of the pre-trained model remain unchanged, reducing computational resources and storage requirements while retaining the rich knowledge learned by the model during pre-training. AdapterFusion~\cite{pfeiffer2020adapterfusion} is an extension of adapter-based fine-tuning that allows the model to learn multiple tasks or adapt to various data distributions simultaneously by fusing multiple adapter modules, with each adapter module focusing on capturing task-specific features. Infused Adapter by Inhibiting and Amplifying Inner Activations (IA3)~\cite{liu2022few} scales the activation layers by injecting learned vectors into the attention and feed-forward modules of the Transformer architecture. Since these learned vectors are only trainable parameters during fine-tuning, IA3 significantly reduces the number of trainable parameters compared to traditional adapter-based fine-tuning, thereby reducing training costs and improving training efficiency. Additionally, IA3 does not introduce inference latency because its adapter weights can be merged with the FM while maintaining the flexibility and adaptability of the model, allowing customized fine-tuning for different tasks and datasets. {\cite{chen2024model} proposes a method called model reprogramming that converts both the input and output domains of an FM by adding two adapter layers, enabling more flexible adaptation. Unlike inserting adapters into each layer in the FM, model reprogramming serves as a more direct way in which only the input and output adapters are inserted upstream and downstream of the FM. During reprogramming, the pre-trained FM is frozen, which means the core of the model still works in the original domain. Such a process reduces the cost of training resources and enables the simultaneous adaptation of a single pre-trained FM to multiple domains. However, because of less trainable adapters, such a method depends more on the similarity between the source domain of the pre-trained FM and the target domain. For example, time-series data can be considered similar to text data since they all persist in a series form \cite{jin2023time}. To better illustrate the ability of model reprogramming, we present an example in Figure~\ref{TS_example}.}
    
    \begin{figure}[!t]
        \centering
        \centerline{\includegraphics[width=1.0\linewidth]{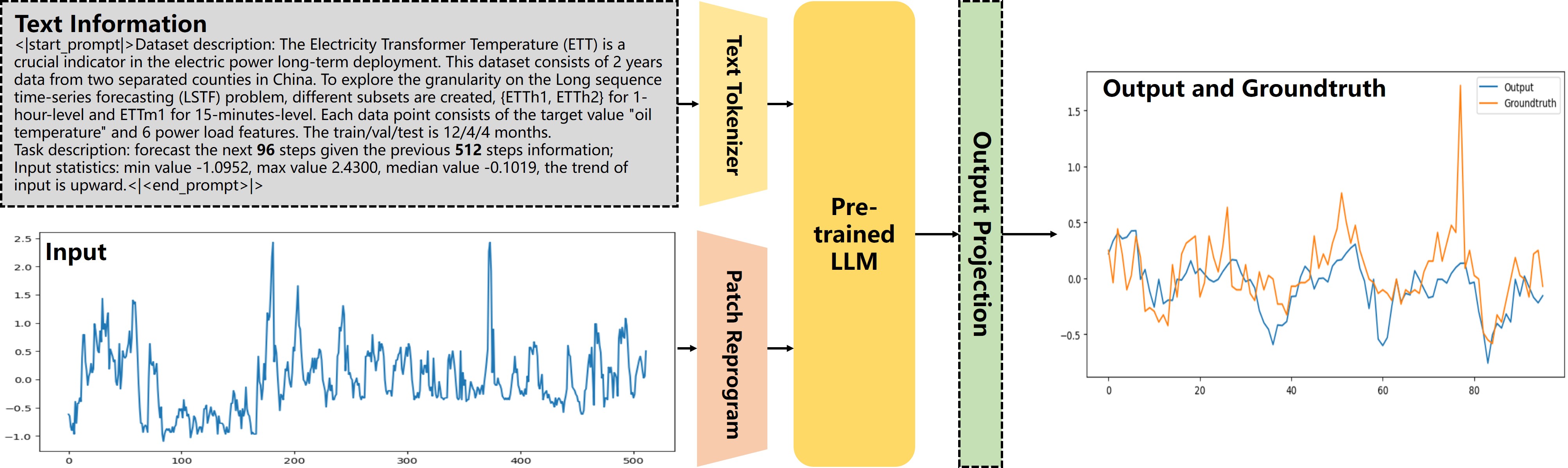}}
        \caption{{An example of model reprogramming. The input time-series data is embedded by a patch reprogram module, which patches the data and then projects each patch embedding into the text space. Besides the time-series part, text information of the dataset is also inputted for LLM to better understand the prediction task. The data used in the example is from \cite{jin2023time}.}}
        \label{TS_example}
    \end{figure}
    
    \item \textbf{Low-rank matrix decomposition based fine-tuning}: Fine-tuning based on low-rank matrix decomposition reduces the number of parameters that need to be updated by decomposing the weight matrices in the pre-trained model into the product of low-rank matrices. Low-rank matrix decomposition can capture the most important information in the weight matrices while keeping the original pre-trained parameters unchanged during fine-tuning, updating only low-rank decomposition matrices, thus reducing the computational and storage requirements during fine-tuning. This method improves fine-tuning efficiency while maintaining or approaching the performance of full-parameter fine-tuning. {In low-rank matrix decomposition-based fine-tuning methods, two low-rank matrices, $A$ and $B$, share the same rank $r$, are introduced to construct an update parameter matrix. The low-rank property of both $A$ and $B$ enables this process to be computationally efficient while still enabling the model to adapt effectively to new tasks.}
    \begin{enumerate}
        \item \textbf{Low-Rank Adaptation (LoRA)}: Low-Rank Adaptation (LoRA)~\cite{hu2021lora} achieves efficient fine-tuning performance by introducing low-rank adaptation matrices to the pre-trained model, rather than updating the original weight matrix directly. LoRA modifies the model by adding a low-rank update to the original weights, as described by the equation:
        \begin{equation}
        W_{\text{new}} = W_{\text{old}} + \Delta W \text{,}
        \end{equation}
        where $W_{\text{old}}$ is the original weight matrix, $\Delta W$ is the low-rank update matrix, and $+$ is the point-wise addition of two matrices with the same shape. The key idea is that $\Delta W$ is constructed as the product of two low-rank matrices $A$ and $B$ , such that:
        \begin{equation}
        \Delta W = A B \text{.}
        \end{equation}
        Here, $A$ and $B$ are low-rank matrices with ranks typically much smaller than that of $W_{\text{old}}$. During fine-tuning, only the matrices $A$ and $B$ are updated, while $W_{\text{old}}$ remains fixed. These low-rank matrices are learned to adapt the model efficiently while keeping the overall number of trainable parameters minimal.
        
        One limitation of LoRA is that it typically applies the same low-rank structure to all layers, ignoring the varying importance of different layers and parameters for downstream tasks. Adaptive Low-Rank Adaptation (AdaLoRA)~\cite{zhang2022adaptive} is an improved method based on LoRA, which adaptively determines which layer parameters need to be updated. By employing adaptive learning rates and task-specific parameter adjustment strategies, AdaLoRA enables the model to automatically adjust the intensity and scope of fine-tuning according to the specific requirements of the task.
        
        Researchers have also found that LoRA's continuous pre-training performance is unsatisfactory on some large-scale datasets. Thus, the Layerwise Importance Sampled AdamW (LISA)~\cite{pan2024lisa} strategy is proposed, where the weight norm distribution of different layers exhibits uncommon skewness. LISA adopts an importance sampling strategy by randomly activating different layers in the FM for optimization. Specifically, LISA consistently updates the bottom embedding layers and the top linear head while randomly updating a small number of intermediate self-attention layers. With memory consumption comparable to LoRA, this method outperforms LoRA and even full-parameter fine-tuning in various downstream fine-tuning tasks.

        \item \textbf{Low-Rank Hadamard Product (LoHa)}: Low-Rank Hadamard Product (LoHa)~\cite{kim2016hadamard} updates the model's weights by introducing the Hadamard product of low-rank matrices. The principle of LoHa can be represented by the following equation:
        \begin{equation}
        W_{\text{new}} = W_{\text{old}} + \Delta W \text{,}
        \end{equation}
        {where $W_{\text{old}}$ is the original weight matrix and $\Delta W$ is the update matrix approximated by a low-rank matrix}. The update matrix $\Delta W$ can be further decomposed into the Hadamard product of two low-rank matrices:
        \begin{equation}
        \Delta W = A \odot B \text{.}
        \end{equation}
        Here, $A$ and $B$ are two low-rank matrices that adjust elements of the original weight matrix by learning key information extracted from the input data, {and $\odot$ is the Hadamard product}. By updating only parameters in $A$ and $B$, LoHa achieves efficient model fine-tuning while maintaining adaptability to new tasks.
        
        \item \textbf{Low-Rank Kronecker Product (LoKr)}: Low-Rank Kronecker Product (LoKr)~\cite{hackbusch2006low} is another parameter-efficient fine-tuning method that emerged after LoHa. LoKr utilizes the properties of the Kronecker product to expand the dimensions of the weight matrix while keeping the increase in the number of parameters within a manageable range. The Kronecker product allows the model to learn complex interactions across different dimensions, particularly useful for capturing high-order relationships in the input data. The updating process of LoKr can be represented as:
        \begin{equation}
        W_{\text{new}} = W_{\text{old}} + \Delta W \text{,}
        \end{equation}
        where $\Delta W$ is the Kronecker product of two low-rank matrices $A$ and $B$:
        \begin{equation}
        \Delta W = A \otimes B \text{.}
        \end{equation}
        {In this formula, two low-rank matrices, $A$ and $B$, are used by the algorithm to compute a large matrix through the Kronecker product (denoted as $\otimes$), which is then updated during the fine-tuning process.} LoKr is particularly suitable for tasks that require increasing the model's dimensions to capture more complex relationships while maintaining similar parameter efficiency to LoHa. However, LoKr may require more complex mathematical operations to handle the Kronecker product, and in some cases, its computational cost may be higher than that of LoHa.
    \end{enumerate}

    \item \textbf{Full fine-tuning}: Full fine-tuning is not constrained by pre-training tasks or data distributions, making it flexible to adapt to various downstream tasks. It allows the model to be directly optimized end-to-end on the data of the final task without the need for additional adapter modules. However, because it requires updating all parameters in the model, full fine-tuning demands significant computational resources and longer training times. Moreover, FMs have a vast number of parameters, and insufficient fine-tuning data may lead to overfitting. Additionally, the intermediate variables generated during full fine-tuning consume a large amount of GPU memory. Therefore, researchers have proposed many parameter-efficient fine-tuning methods mentioned earlier, which can reduce resource consumption and training time while maintaining performance~\cite{ding2023parameter}.
\end{enumerate}

\subsection{Customization of the foundation model based on pre-trained modules}

FMs may contain millions or even billions of parameters. Through domain-specific customization, it's possible to reduce parts of the model that need training, thereby significantly lowering training costs. This approach is known as the customization of the FM based on pre-trained modules. Such customization means utilizing the knowledge embedded in model parameters during the pre-training process when constructing a new model. As mentioned earlier, in the architecture of FMs, there are typically five main modules: MEs, IPs, BCs, OPs and MDs. Among them, the ME, BC, and MD carry a large amount of knowledge as they are directly involved in data encoding, processing, and decoding. In contrast, the IP and OP themselves carry less model knowledge. In some cases, it is unnecessary to explicitly train these modules. Therefore, when customizing FMs, the IP and OP are often excluded.

Next, this paper will detail how to customize FMs based on the pre-trained ME, BC, and MD. Through this approach, we can effectively leverage pre-trained models' knowledge while reducing the demand for computational resources, making the model more suitable for specific tasks and environments.

\subsubsection{{Customization of modality encoder}}

General-purpose FMs often adapt to the distribution characteristics of a vast dataset during training, internalizing ample domain knowledge in their model parameters. They are thus well-suited as feature extraction modules for customized FMs. By utilizing the pre-trained model's pre-existing feature extraction module as the ME for domain data, downstream task modules can be seamlessly integrated after this module to fulfill task requirements. The ME encapsulates knowledge about crucial features of the data. There are two main approaches to customizing MEs:

\begin{figure}[!t]
    \centering
    \centerline{\includegraphics[width=1.0\linewidth]{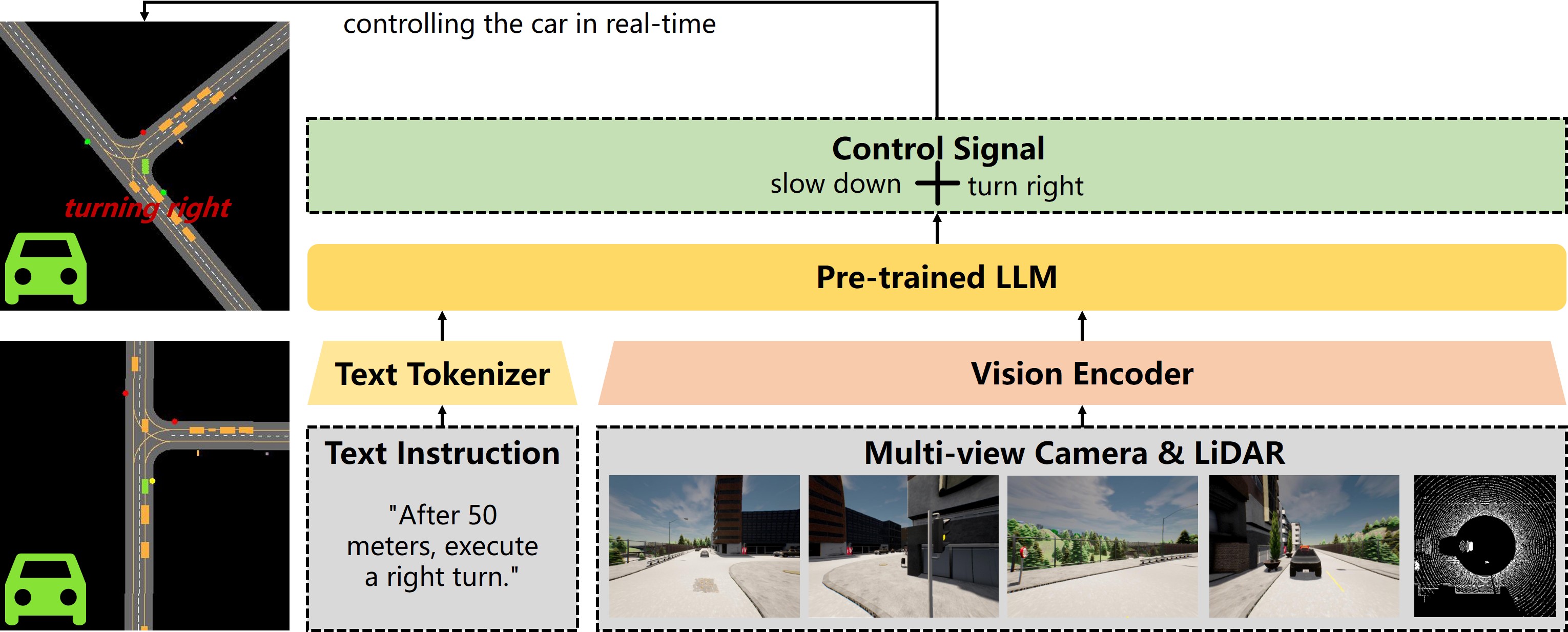}}
    \caption{{An example of using LLM as the pre-trained BC to adopt to the autonomous driving area. In this example, multi-view camera data and LiDAR data are sent into a vision encoder aligned with LLM's feature space. By incorporating the user's instructions and environmental information, the LLM can generate real-time control signals for the car. The data used in the example is from~\cite{shao2024lmdrive}.}}
    \label{autodrive_example}
\end{figure}

\begin{itemize}
    \item \textbf{Customization within the same modality across different data domains}: This approach involves transferring the knowledge of the ME trained on the source data domain to the target data domain. It typically entails aligning the feature distribution of the source and target domain data. Fine-tuning the pre-trained ME from the source domain on the target domain enables it to adapt to the characteristics of the new data. Specifically, this can be achieved by adjusting or adding layers to the encoder. Additionally, domain adaptation techniques such as domain adversarial training or domain-invariant feature extraction techniques can reduce the distribution discrepancy between the source and target domains.
    
    \item {\textbf{Customization across modalities}: When it comes to multi-modality FMs, there are often cases that lack corresponding pre-trained encoders for certain modalities. In such situations, a cross-modality customization strategy can be utilized, adapting encoders to process new data modalities similar to the original modality. Although the modality is different, the source and target modality with the same data structure, such as natural images and thermal images, can be simultaneously fed into one ME. Such customization involves fine-tuning a pre-trained ME using task-specific datasets. For instance, ImageBind treats depth and thermal imaging data as single-channel images, using image encoders to extract features for thermal data. Initializing the model with weights pre-trained on image datasets can lead to faster convergence than random initialization and enhance generalization. This strategy is particularly useful in applications like autonomous driving, where vehicles need to process diverse data modalities, including visual, depth, thermal, and LiDAR (Light Detection And Ranging) data. To more directly understand this process, we present another example in Figure~\ref{autodrive_example}. In this example, natural images captured by multi-view cameras are fed into the pre-trained vision encoder with LiDAR data.}

    By adapting image encoders to handle thermal imaging, the model can better understand and process thermal data, improving the overall performance of the autonomous driving system. 
    This cross-modal customization increases the flexibility of multi-modal systems and enhances their adaptability and robustness, providing a powerful tool for solving complex problems.

\end{itemize}

\subsubsection{{Customization of backbone calculator}}

For multi-modality FMs, the BC is the core computational component for processing encoded feature vectors and performing tasks such as classification and generation. The BC of customized FMs can be customized from pre-trained models to leverage the complex feature processing and task execution capabilities learned from large-scale datasets. This approach avoids training the BC from scratch but requires constructing appropriate previous modules to encode the data into feature vectors that the BC can process. For example, NExT-GPT~\cite{wu2023next} converts raw data from various modalities into language modality feature vectors before feeding them into a pre-trained LLM, allowing the model to process inputted tokens according to task requirements. There are two main approaches to customize pre-trained BCs:

\begin{itemize}
    \item \textbf{Customization from a single pre-trained BC}: Utilizing a general-purpose FM (e.g., LLaMA) as the BC to process data from the central modality. Customizing a pre-trained BC to a specific domain application typically requires fine-tuning to adapt to domain-specific data characteristics and task requirements. This step can be performed on a limited domain dataset, fine-tuning the model parameters to optimize its processing capabilities for the domain-specific data.

    \item \textbf{Modular combination of multiple pre-trained BCs}: Modular combination is a flexible design method in deep learning architectures that allows integrating multiple specialized pre-trained models into a unified framework based on task requirements. The Mixture of Experts (MoE) model~\cite{ma2018modeling} can serve as an effective mechanism to further optimize this modular combination. The MoE model introduces multiple expert networks and uses a gating mechanism and mixing strategy to dynamically select and combine the outputs of these experts, enabling specialized processing for different tasks or data subsets. \textcolor{black}{Recent advancements in large-scale foundation models have adopted sparse MoE architectures to achieve a favorable balance between model capacity and computational efficiency. For instance, DeepSeek-V3 employs a large-scale MoE setup with a total of 671 billion parameters, yet activates only 37 billion parameters per token using a top-2 expert routing strategy~\cite{liu2024deepseek}. It introduces an efficient expert-balancing mechanism without auxiliary loss, and leverages a multi-token prediction objective for improved generalization. Similarly, Mixtral-8x7B~\cite{jiang2024mixtral} activates only two out of eight expert subnetworks (each a 7B model) per forward pass, enabling performance comparable to much larger dense models while significantly reducing inference costs. These examples demonstrate how sparse MoE designs have become core architectural elements in state-of-the-art foundation models, offering scalable specialization with manageable computational overhead.}
    
    The primary function of the gating mechanism is to determine how input data should be distributed among experts. It generates a weight or score for each expert based on the characteristics of the input data, reflecting each expert's capability or suitability for processing the current input. The output of the gating mechanism is typically used to guide the mixing strategy, indicating the importance of each expert for the current input. The mixing strategy then combines the outputs of multiple experts according to specific rules to generate the final model output. This strategy can be simple, such as averaging or weighted averaging, or more complex, involving probability distributions of model outputs or other advanced methods.
    
    For example, for complex tasks requiring both image recognition and language understanding, one expert network might excel at identifying edges in images. In contrast, another expert network might be adept at understanding semantic relationships in natural language. The MoE model's gating mechanism can automatically adjust each expert network's participation level based on the input data's characteristics and the task requirement, allowing the model to flexibly invoke the most appropriate expert network when dealing with mixed visual and language inputs, achieving optimal performance. Moreover, MoE models are highly scalable. They can adapt to new task requirements or data types by adding new expert networks and updating the gating mechanism, making them suitable for constructing flexible customized FMs.
\end{itemize}

\subsubsection{{Customization of modality decoder}}

MDs play a crucial role in multi-modality foundation pre-trained models for converting processed feature vectors back into the form of the original data. In generative tasks, such as converting text to images or audio to text, MDs need to accurately decode the feature vectors to reconstruct understandable original data and exhibit a certain level of creativity. Some pre-trained MDs can also understand and process multi-modality feature inputs. For instance, CoDi-2 can utilize both text and audio as conditions to control image generation. By customizing such pre-trained decoders, there is no need to train complex decoder structures from scratch, allowing them to be directly applied to image generation tasks.

Here are methods to effectively utilize pre-trained MDs for customization:

\begin{enumerate}
    \item \textbf{Fine-tuning pre-trained MDs}: Similar to MEs, MDs can be fine-tuned on a specific task's dataset to adapt to new task requirements. This process usually involves adjusting the last few layers of the decoder or adding new layers to better capture the data characteristics of the specific domain.

    \item \textbf{Customizing cross-modality generative MDs}: In cross-modality generation tasks, pre-trained MDs can be directly used to generate data in the target modality. The conditional information is first encoded into feature vectors through a conditional encoder and then combined with feature vectors of the original data to achieve conditional generation. The prerequisite for this functionality is ensuring that the decoder can correctly understand input feature vectors, which may involve adjustments to the BC and OP.
\end{enumerate}

\subsection{Construction of the foundation model without pre-trained modules}

When it is not possible to construct an FM through transferring from pre-trained models, it becomes necessary to design and train the corresponding modules. We will first provide a general analysis of the architectures of single-modality and multi-modality FMs as the foundation for constructing each component.

\textbf{Single-modality FMs} consist of three core modules: the ME, the BC, and the MD. For example, in LLaMA-2~\cite{touvron2023llama2}, the ME and decoder are specifically designed for the language modality, utilizing the byte pair encoding (BPE) algorithm for encoding and decoding functions. And the BC is a massive autoregressive Transformer model. In this way, LLaMA-2 achieves a complete processing workflow of "input raw text -- input text feature vectors -- output text feature vectors -- output raw text." Additionally, Bai et al.~\cite{bai2024sequential} introduce the concept of visual sentences and propose a large visual model (LVM) that can autoregressively generate the required image based on visual sentences. It realizes in-context learning within the pure image modality, which enables the model to infer tasks directly from image modality prompts and generate corresponding results. This not only explores the potential of pure visual input but also provides a new perspective for constructing domain-specific FMs --- the central modality does not have to be limited to language but any modality widely used in a specific domain.

\textbf{Multi-modality FMs} require the additional IPs and OPs to achieve modality alignment. For instance, CoDi-2~\cite{tang2024codi} first utilizes multiple MEs proposed in ImageBind~\cite{girdhar2023imagebind} to process input data, aligning all corresponding modalities to the image modality. Then, the feature vectors of the image modality are transformed into the feature space of the language modality through a MLP. In detail, it uses the pre-trained autoregressive Transformer of the LLM LLaMA-2-7b-chat-hf as the BC's foundation. The image and audio features processed by the BC are converted back to the image domain via two MLPs. They are then used as control vector input of a Diffusion-based generative model to obtain the final image and text results. The training losses include text generation, modality conversion, and data generation loss. So that the multi-modality feature processing capability of the BC and the modality conversion capability of the two MLPs can be trained simultaneously in an end-to-end way. The model's modality alignment is reflected in two terms. On the one hand, the model aligns the feature vectors of multiple modalities to the imaging modality through the pre-trained MEs of ImageBind. On the other hand, it also converts between image feature vectors and text feature vectors via the MLP.

In summary, constructing an FM begins with determining the data modalities and selecting the central modality. Next, the ME and IP are implemented to convert raw data from different modalities into central modality feature vectors, which the BC will then process. Following this, OP and MD are designed to convert the feature vectors from the BC back into the original data forms of each modality. Once the model structure is constructed, the training process can begin. In the following, we will provide a detailed introduction to each module's implementation principles and construction methods.

\subsubsection{Constructing modality encoder}

Constructing a ME means designing a neural network capable of extracting feature vectors from data. The general steps for building a ME are as follows:

\begin{enumerate}
    \item \textbf{Preprocessing into suitable data structures}: Choose an appropriate data structure based on the characteristics of the data modality for subsequent model usage. For example, in audio processing, a common approach is to convert time-domain signals into spectrograms and then use a neural network designed for images to extract features. Another method is to view audio as sequential vectors and use neural networks for sequences, like time series or neural language, to process it. When selecting the target data structure, researchers need to balance task requirements and processing difficulty, ensuring that the data structure represents domain knowledge sufficiently and is suitable for downstream model processing. Additionally, since domain-specific FMs need to be functionally versatile, the compatibility of various task inputs should be considered when choosing the target data structure.

    \item \textbf{Designing the network architecture}: Design the network architecture according to the characteristics of the input data structure. For example, Transformer architectures can be used for text data to capture long-range dependencies, while CNN-based or ViT-based models can be employed for image data to extract features.

    \item \textbf{Training the ME}: Pre-train the ME using a dataset with sufficient quantity and variety of samples to learn the general features and distributions of the modality data. Pre-training is the process of infusing the model with knowledge. If the dataset size or diversity is inadequate, the model might not learn a complete representation of the modality data. One method to train an ME is to combine the ME and MD into an autoencoder and perform unsupervised training by minimizing reconstruction error. Another training method is designing and training a model for a specific task by supervised training, then transferring the model's upstream part as the ME. However, this method cannot get a compatible MD, which might affect the design and functionality of subsequent modules. Therefore, when designing the ME, it is essential to consider the consistency of the entire FM architecture.
\end{enumerate}

\subsubsection{Constructing input projector}

The role of IPs is to project data from different modalities into a common feature space. As discussed in section \ref{Model Training}, modality alignment can be achieved through either fusion encoders or dual encoders. When constructing IPs, the key decision is whether to use a bridging strategy to integrate input vectors from different modalities or to use a fine-tuning approach to bring the projectors of different modalities closer together. These two strategies correspond to the concepts of fusion encoders and dual encoders, respectively. During training, using loss functions from multi-modality understanding tasks, such as multi-modality classification or generation task losses, can train the model's cross-modality projection capability. Additionally, an end-to-end training approach can optimize the overall performance of the FM and train cross-modality projections at the same time. As previously mentioned, the CoDi-2 model~\cite{tang2024codi} utilizes ImageBind~\cite{girdhar2023imagebind} encoders aligned by CLIP as the image and audio MEs and part of the IP. It then incorporates an MLP as another part of the IP. During the end-to-end training process of the FM, the MLP is optimized to align the image and audio to text.

\subsubsection{Constructing backbone calculator}

The BC is capable of understanding and generating feature vectors of the central modality. Constructing a BC for a specific domain begins with identifying the most common and information-rich data modality in that domain as the processed modality by the BC. The model architecture is then designed based on this modality. Currently, mainstream model architectures are based on Transformers. A complete Transformer model consists of an encoder and a decoder, where the encoder analyzes the input data to extract compact feature representations, and the decoder uses these feature representations to generate the output content. Since the structures of the encoder and decoder are different, generally, the encoder has stronger comprehension capabilities, while the decoder has more powerful generative capabilities. Transformer-based FM BCs have three main architectural forms: encoder-only architectures, decoder-only architectures, and encoder-decoder architectures. The characteristics of these three architectures are summarized in Table~\ref{tab3}.

\begin{table}[!t]
    \footnotesize
    \caption{Comparison between encoder-only structure, decoder-only structure, and encoder-decoder structure}
    \label{tab3}
    \tabcolsep 17.5pt
    
    \begin{tabular}{ccccc}
    \toprule
    Model Architecture &  \makecell[c]{Generative \\ Capability} & \makecell[c]{Understanding \\ Capability} & Computation & Examples \\
    \hline
    \makecell[c]{Encoder-only} & \makecell[c]{Low} & \makecell[c]{High} & Low & BERT~\cite{devlin2018bert} \\ \hline
    {Decoder-only} & {High} & {Low} & {Low} & \makecell[c]{GPT Series~\cite{radford2018improving, radford2019language, brown2020language, achiam2023gpt}, \\LLaMA Series~\cite{touvron2023llama, touvron2023llama2}} \\ \hline
    {Encoder-decoder} & {High} & {High} & {High} &\makecell[c]{BART~\cite{lewis2019bart}, T5~\cite{raffel2020exploring}} \\ 
    \bottomrule
    \end{tabular}
\end{table}

\begin{enumerate}
    \item \textbf{BC based on encoder-only architecture}: The encoder-only architecture consists solely of the encoder part of the Transformer. It is typically used for tasks that require understanding input text rather than generating new text sequences, such as text classification and sentiment analysis. Due to containing only the encoder part, the encoder model structure is relatively simple but can only produce fixed-length outputs, resulting in limited generation capability. Regarding generation tasks, BCs based on an encoder-only model can only handle tasks such as sequence completion. BERT~\cite{devlin2018bert} is a well-known example of an encoder-only model.
    \item \textbf{BC based on decoder-only architecture}: In the decoder-only architecture, the decoder directly processes the input sequence and generates the output sequence without a separate encoder to compactly represent the input sequence. This reduces parameter count and computational overhead but also makes it more challenging for the model to understand input sequences, thus limiting its ability to handle long sequences. This architecture does not require explicit context representation when generating output sequences but captures information automatically within the sequence through self-attention mechanisms. BCs based on decoder-only architecture typically employ autoregressive generation, meaning they generate words or characters one by one based on previously generated content to complete sequence text generation tasks. Models like the GPT series~\cite{radford2018improving, radford2019language, brown2020language, achiam2023gpt} and the LLaMA series~\cite{touvron2023llama, touvron2023llama2} of foundation language models belong to the decoder-only architecture.
    \item \textbf{BC based on encoder-decoder architecture}: The encoder-decoder architecture can simultaneously possess the understanding capability of an encoder and the generation capability of a decoder, but also results in higher parameter count and computational costs. This architecture is adopted by models like Meta's BART~\cite{lewis2019bart} and Google's T5~\cite{raffel2020exploring}.
\end{enumerate}

\subsubsection{Constructing output projector and modality decoder}

\begin{figure}[t!]
    \centering
    \begin{minipage}[c]{0.4\textwidth}
        \centering
        \includegraphics[height=2.7cm]{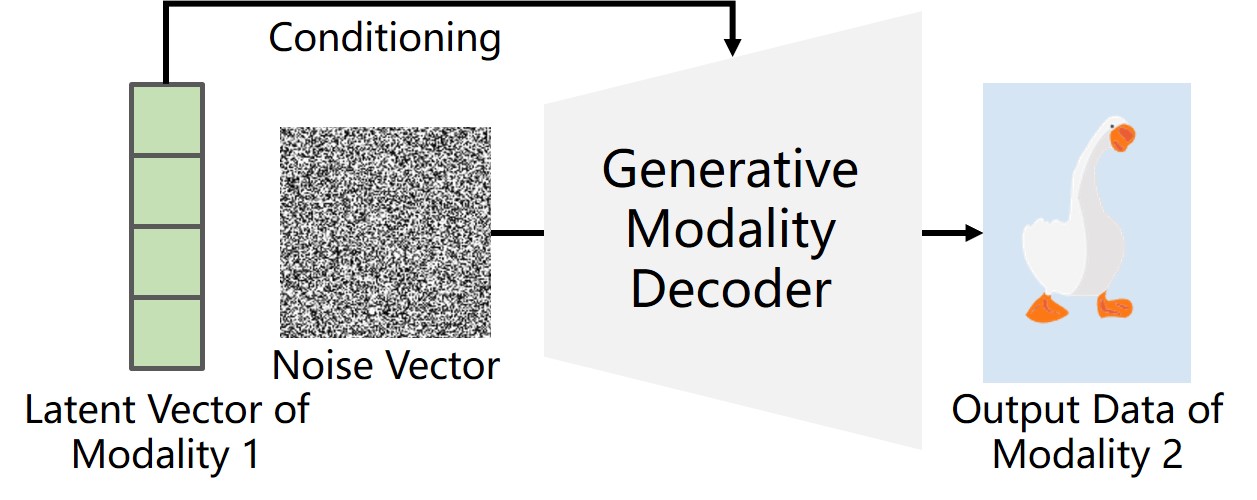}
        
        (a)
    \end{minipage}
    \hfill
    \begin{minipage}[c]{0.55\textwidth}
        \centering
        \includegraphics[height=2.7cm]{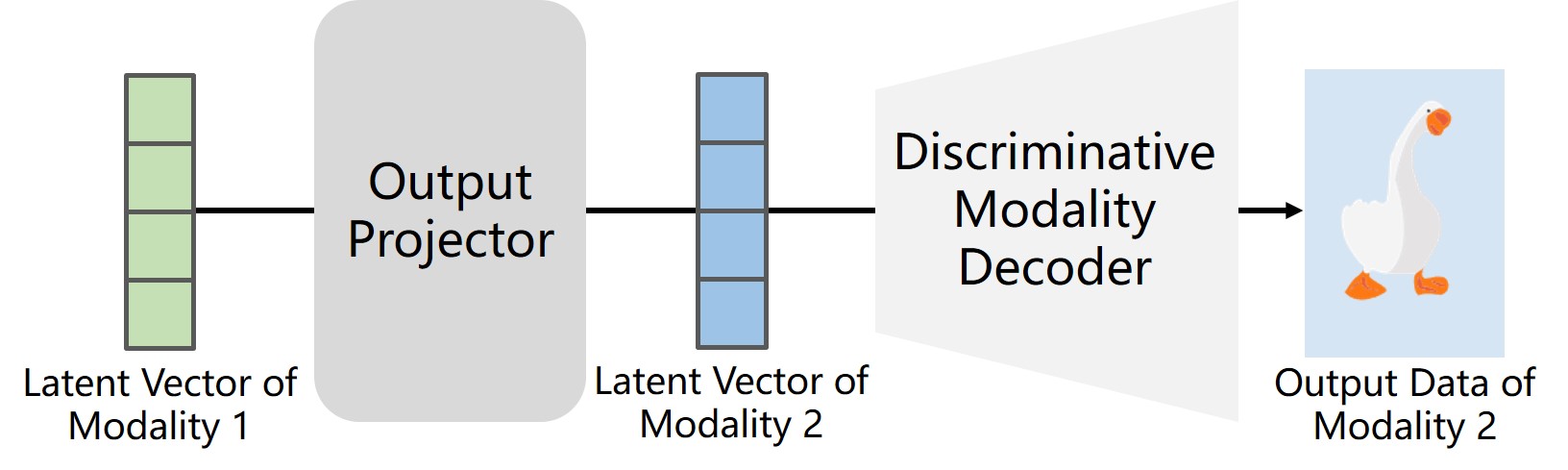}
        
        (b)
    \end{minipage}
    
    \caption{The workflow of OP and MD. (a) Generative MD;(b) Discriminative MD.}
    \label{fig7}
\end{figure}

MDs come in two types: generative and discriminative. Generative MDs can produce high-quality data samples given conditional information. Discriminative MDs, on the other hand, excel in the precise reconstruction of data samples based on input vectors. The type of MDs also affects the design of OPs. Thus it requires a joint consideration of these two modules. Figure~\ref{fig7} (a) and (b) respectively illustrate the operational processes of generative and discriminative MDs.

\begin{enumerate}
    \item \textbf{Generative MD}: A generative MD utilizes a generative model as its neural network, which can control the generation process using conditional information. It takes other modality features as conditional information and generates data as decoding output. Such a generative model can be termed as a generative MD without the need for explicit construction of OPs. For instance, DiT~\cite{peebles2023scalable}, a diffusion-based image generation model, can progressively generate images based on previous results and conditional vectors, where cross-attention mechanisms serve the function of the OP. Similarly, the VAR model~\cite{tian2024visual} also uses the self-attention mechanism as the OP function. By incorporating modality labels in the starting position, VAR knows which content needs to be controlled for generation and generates images in an autoregressive way.
    % By incorporating modality labels in the contextual input, VAR, using an autoregressive generation mechanism, knows which content needs to be controlled for generation, with the rest being previously generated content, thereby achieving autoregressive image generation.
    
    Generative MDs typically employ end-to-end training strategies. During training, the generation part of the model and the modality interaction part are optimized simultaneously. The training objective usually involves minimizing the difference between generated data and real data while ensuring that generated data meets given conditions. For example, if the model's objective is to generate images based on textual descriptions, a large number of text-image pairs are used during training. The model is optimized by comparing the similarity between generated images and real images. This similarity can be measured using pixel-level loss functions (such as mean squared error) or more advanced perceptual losses (such as VGG loss). Additionally, adversarial training can be employed to enhance generation quality.
    
    \item \textbf{Discriminative MD}: A discriminative MD is only for directly reconstructing feature vectors into their original data form. Thus, it requires an explicit OP to convert feature vectors from other modalities to the target modality for use. For example, when using the decoder of VQGAN~\cite{esser2021taming} as the discriminative MD in a multi-modality FM, it is necessary to first explicitly construct an OP to transform feature vectors from other modality domains to the image modality, and then use the decoder portion to decode the feature vectors into their original data form. This explicit OP is typically trained using a supervised way, where the model takes feature vectors from other modalities as input and learns to project these features into feature vectors of the target modality by minimizing the error between the two feature vectors. Besides, MDs are trained together with MEs as autoencoders, aiming to improve reconstruction performance by minimizing reconstruction errors.
\end{enumerate}

\section{Applications of domain-specific foundation models}

FMs have emerged as powerful tools with extensive application prospects across various domains. These FMs possess the capability not only to handle massive data and complex tasks but also to bring about new breakthroughs and innovations.

\begin{itemize}
    \item \textbf{Telecommunications}: FMs are expected to be widely applied in sensing, transmission, network performance optimization, etc. For example, an LLM fine-tuned for the telecommunications domain can be used to process network log data, model, and solve specific network problems~\cite{bariah2023large, cui2025overview}. Additionally, FMs in telecommunications, through the utilization of spatio-temporal correlations and knowledge reasoning~\cite{zhou2024large}, are expected to identify and prevent experience issues caused by degraded service quality and delayed fault response times. Recent works also successfully incorporate telecommunication standards into LLMs' reasoning~\cite{maatouk2023teleqna, chen2025first}. As a pioneering work on how to introduce knowledge of the wireless communication field to LLMs, \cite{shao2024wirelessllm} presents a comprehensive framework that elaborates on the principles and methods for constructing a ``WirelessLLM" and provides examples to demonstrate its functionality. Specifically, TelecomGPT\cite{zou2024telecomgpt} acquires domain knowledge in telecom-specific datasets through continual pre-training, instruction tuning, and alignment tuning, after which it is able to answer related questions and generate codes in the telecom domain. This lays the foundation for fine-grained network optimization in real time. We illustrate the advantages of domain-specific enhancement by presenting the results from~\cite{zou2024telecomgpt}, as summarized in Table~\ref{table:telecomgpt}. This table highlights the performance of LLMs that have undergone domain-specific enhancements, denoted by TP (continual pretraining), TI (instruct tuning), and TA (alignment tuning) in their names. It is evident that these enhancements lead to significant improvements in performance across various metrics. The efficiency of foundation models is crucial in use~\cite{bommasani2023holistic}. In response to the significant challenges in data and communication faced when training FMs, \cite{chen2024role} proposes multiple new paradigms to explore how Federated Learning can be utilized in a wireless context to address these challenges. In industrial scenarios, the business understanding capability of FMs is also expected to help optimize signal transmission and scheduling strategies, improving network efficiency. In research on network FMs, the NetGPT architecture proposed in the article~\cite{chen2024netgpt} is expected to become an effective approach to achieve endogenous intelligence in telecommunications networks, while the article~\cite{tong2023ten} discusses the challenges and issues that may be encountered in constructing FMs for telecommunications. The NetLLM proposed in the article~\cite{wu2024large} adapts LLM to serve several specific downstream communication tasks. \cite{xie2024towards} introduces a new memory mechanism to facilitate the embedding of FMs into the semantic communication process. By leveraging the FMs' capability to understand and generate multimodal data, it becomes feasible to propose new and more complex transmission approaches that enhance transmission performance. In telecommunications, training domain-specific FMs on large datasets from telecommunication systems involves significant costs for data storage, processing power, and specialized hardware. Deployment costs are often high due to the need for real-time processing capabilities, especially in edge environments~\cite{you2025ai}. Additionally, ongoing operational costs arise from continuous monitoring, model updating, and adaptation to evolving network conditions.

    \begin{table}[ht]
        \footnotesize
        \caption{Performance comparison of Telecom multiple-choice question (MCQ) answering on TeleQnA~\cite{maatouk2023teleqna} dataset of mainstream LLMs and different telecom-specific LLMs obtained during different stages of TelecomGPT's training pipeline}
        \label{table:telecomgpt}
        % \tabcolsep 17.5pt
        % \centering
        \def\tabblank{\hspace*{10mm}}
        \begin{adjustbox}{width=\textwidth}
        \begin{tabular}{lcccccc}
            \toprule
            LLMs & Lexicon & Research Overview & Research Publications & Standards Overview & Standards Specifications & Avg \\
            \midrule
            % GPT-4o & 92 & \textbf{81.73} & \textbf{79.54} & \textbf{83.87} & \textbf{62.77} & \textbf{78} \\
            % GPT-4 & 92 & 77 & 78 & 79 & 60 & 75 \\
            % GPT-3.5 & \textbf{96} & 66.35 & 66.98 & 64.52 & 56.38 & 66 \\
            % \hline
            Llama3-8B & 72 & 51.92 & 65.11 & 56.45 & 36.17 & 56.20 \\
            Llama3-8B-Instruct & 80 & 67.31 & 69.77 & 59.68 & 50 & 64.80 \\
            Llama3-8B-TI & \textbf{96} & 69.23 & \textbf{74.88} & \textbf{74.19} & 56.38 & \textbf{71.20} \\
            Llama3-8B-TI-TA & 92 & \textbf{73.08} & 71.63 & 72.58 & \textbf{58.51} & 70.60 \\
            \hline
            Mistral-7B & 72 & 49.04 & 51.16 & 50 & 34.04 & 48.40 \\
            Mistral-7B-Instruct & \textbf{84} & 64 & 65 & 56 & 51 & 62 \\
            Mistral-7B-TI & \textbf{84} & 67.3 & 70.69 & 56.45 & \textbf{51.06} & \textbf{65.2} \\
            Mistral-7B-TI-TA & \textbf{84} & \textbf{70.19} & \textbf{73.95} & \textbf{61.29} & 48.94 & 64 \\
            \hline
            LlaMA-2-7B & 62.5 & 52.24 & 49.18 & 48.28 & 40 & 48.94 \\
            LlaMA-2-7B-TI & \textbf{84} & 57.69 & 63.26 & 56.45 & \textbf{50} & 59.80 \\
            LlaMA-2-7B-TP-TI & 81.82 & \textbf{63.92} & \textbf{67} & \textbf{70} & 47.48 & \textbf{63.79} \\
            \bottomrule
        \end{tabular}
        \end{adjustbox}
    \end{table}
    
    \item \textbf{Autonomous driving}: FMs play a core role in various key aspects such as vehicle perception, decision-making, and motion control~\cite{cui2024survey}. Specifically, perception tasks in autonomous driving involve real-time monitoring of the vehicle's surroundings, including other vehicles, pedestrians, traffic signs, and road conditions~\cite{wang2023drivemlm}. FMs, by analyzing data collected from sensors such as cameras, radar, and LiDAR, can identify various objects and construct detailed maps of the vehicle's surroundings. This advanced perception capability is fundamental to achieving safe autonomous driving. At the decision-making level, FMs need to make rapid and accurate judgments based on perceived information, such as avoiding obstacles, selecting appropriate driving paths, and devising optimal driving strategies in complex traffic situations~\cite{cui2023drivellm, fu2024drive}. Multi-modality FMs like DriveGPT~\cite{xu2023drivegpt4} can not only process visual data but also understand and respond to language-mode instructions, such as planning routes based on voice input destinations. Additionally, pFedLVM~\cite{kou2024pfedlvm} can utilize the powerful performance of pre-trained visual FMs for image feature extraction, serving as the basis for downstream tasks. Considering the scarcity of high-quality public datasets for training LLMs in autonomous driving scenarios, {\cite{chen2024empowering} employs federated instruction tuning and expands the dataset by generating new instruction tracking data to mitigate the data scarcity. DriveMLM~\cite{wang2023drivemlm} is an LLM-based framework that can perform close-loop autonomous driving. It uses driving rules, user commands, and sensor inputs as its multi-modality LLM core inputs to model behavior planning. It standardizes decision states to connect language decisions with vehicle control commands.}
    {Training domain-specific FMs for autonomous driving requires extensive labeled data collection from real-world scenarios, which is costly and time-consuming. Additionally, the training demands substantial computational resources, particularly with large sensor data like LiDAR, radar, and camera feeds. Deploying these models involves significant expenses for hardware integration, safety certifications, and real-time vehicle processing.}
    
    \item \textbf{Mathematics}: Since many general-purpose LLMs are typically pre-trained on datasets that include open-source mathematical corpora, they inherently possess some capacity to address mathematical problems, and through domain-specific enhancements, this capacity can be significantly improved~\cite{ahn2024large, yang2023gpt, yue2023mammoth, wang2023mathcoder, wu2024mathchat}. {The MAmmoTH proposed in~\cite{yue2023mammoth} combined chain-of-thought and program-of-thought, fully leveraging the understanding capability of large language models and the computational power of programming languages, achieving good performance in mathematical reasoning. To provide a clear demonstration of the benefits of domain-specific enhancements, we collect the results from~\cite{yue2023mammoth} in Table~\ref{tab:mammoth}. MAmmoTH-Coder shows a clear performance advantage with general-purpose open-source LLMs and is comparable with closed-source LLMs.} \cite{wang2023mathcoder} indicates through experiments that the seamless integration of LLMs' programming and reasoning abilities enables them to model and solve complex problems progressively.
    {Due to the inherent complexity of mathematical problems and the need for vast, specialized datasets, training domain-specific foundation models in mathematics for tasks such as theorem proving or algorithm optimization can be challenging.}
    
    \begin{table}[ht]
        \footnotesize
        \centering
        \small
        \caption{Experimental results on mathematical evaluation datasets including math problems from elementary, high school, and college levels. Math-SFT? means whether the model has been instruction-tuned on any math reasoning datasets}
        \label{tab:mammoth}
        \def\tabblank{\hspace*{10mm}}
        \begin{adjustbox}{width=\textwidth}
        \begin{tabular}{lcccccc|c}
            \toprule
            Model & Base & Math-SFT? & GSM8K & MATH & AQuA & NumGLUE & Avg \\
            \midrule
            % \multicolumn{8}{c}{Closed-source Model} \\
            % \hline
            % GPT-4 & - & Unknown & 92.0 & 42.5 & 72.6 & 74.7 & 70.5 \\
            % GPT-4 (Code-Interpreter) & - & Unknown & 97.0 & 69.7 & - & - & - \\
            % PaLM-2 & - & Unknown & 80.7 & 34.3 & 64.1 & - & - \\
            % Claude-2 & - & Unknown & 85.2 & 32.5 & 60.9 & - & - \\
            % Codex (PoT) & - & No & 71.6 & 36.8 & 54.1 & - & - \\
            % ART (InstructGPT) & - & Unknown & 71.0 & - & 54.2 & - & - \\
            % \midrule
            % \multicolumn{8}{c}{7B Parameter Model} \\
            % \hline
            Llama-1 & - & No & 10.7 & 2.9 & 22.6 & 24.7 & 15.5 \\
            Llama-2 & - & No & 14.6 & 2.5 & 30.3 & 29.9 & 19.3 \\
            Galactica-6.7B & GAL & GAL-Instruct & 10.2 & 2.2 & 25.6 & 25.8 & 15.9 \\
            Code-Llama (PoT) & - & No & 25.2 & 13.0 & 24.0 & 26.8 & 22.2 \\
            AQuA-SFT & Llama-2 & AQuA & 11.2 & 3.6 & 35.6 & 12.2 & 15.6 \\
            Llama-1 RFT & Llama-1 & GSM8K & 46.5 & 5.2 & 18.8 & 21.1 & 22.9 \\
            WizardMath & Llama-2 & GSM8K+MATH & 54.9 & 10.7 & 26.3 & 36.1 & 32.0 \\
            MAmmoTH & Llama-2 & MathInstruct & 53.6 & 31.5 & 44.5 & 61.2 & 47.7 \\
            MAmmoTH-Coder & Code-Llama & MathInstruct & \textbf{59.4} & \textbf{33.4} & \textbf{47.2} & \textbf{66.4} & \textbf{51.6} \\
            \bottomrule
        \end{tabular}
        \end{adjustbox}
    \end{table}

    \item \textbf{Medicine}: The applications of FMs in medicine encompass various aspects, including disease diagnosis, patient treatment, analysis and prediction of genetic and protein structure data, and medical education~\cite{thirunavukarasu2023large, garg2023exploring, xi2023rise, zhang2023huatuogpt, zhang2023biomedgpt, kazerouni2023diffusion}. {The HuatuoGPT proposed in the article~\cite{zhang2023huatuogpt} can simulate the diagnosis and treatment process of doctors, providing preliminary medical consultation and advice for patients. This model not only reduces the workload of doctors but also enables patients to receive timely medical services in remote or resource-limited environments. To illustrate the advantages of domain-specific enhancements effectively, we present part of the experimental results from \cite{zhang2023huatuogpt} in Table \ref{tab:huatuogpt}. When contrasted with LLMs without medical domain-specific enhancements, HuatuoGPT demonstrates distinct comparative advantages.} \cite{zhang2023biomedgpt} proposes BiomedGPT, which is capable of performing multimodal medical tasks and has achieved excellent performance in understanding and generating medical-related content.
    {In the medical area, the training datasets are often expensive to acquire due to the stringent requirements for privacy and regulatory compliance. Moreover, it is expensive to pay medical experts to annotate data.}

    \begin{table}[ht]
        \centering
        \caption{Experimental results on Chinese medical QA dataset~\cite{li2023huatuo}}
        \label{tab:huatuogpt}
        
        \def\tabblank{\hspace*{10mm}}
        \begin{adjustbox}{width=\textwidth}
        \begin{tabular}{lccccccccccc}
            \toprule
            Dataset & Model & BLEU-1 & BLEU-2 & BLEU-3 & BLEU-4 & GLEU & ROUGE-1 & ROUGE-2 & ROUGE-L & Distinct-1 & Distinct-2 \\
            \midrule
            \multirow{6}{*}{cMedQA2} 
            & T5 (fine-tuned) & 20.88 & 11.87 & 7.69 & 5.09 & 7.62 & 27.16 & 9.30 & 20.11 & 0.41 & 0.52 \\
            % \hline
            \cline{2-12}
            & DoctorGLM & 13.51 & 7.10 & 3.72 & 2.00 & 5.11 & 22.78 & 5.68 & 12.22 & \textbf{0.85} & 0.96 \\
            & ChatGPT & 19.21 & 7.43 & 3.14 & 1.24 & 5.06 & 20.13 & 3.10 & 12.57 & 0.69 & \textbf{0.99} \\
            & ChatGLM-6B & 24.90 & 12.74 & 6.99 & 3.87 & 8.49 & 28.52 & 7.19 & \textbf{18.21} & 0.68 & \textbf{0.99} \\
            & Ziya-LLaMA-13B & 27.03 & 13.87 & 7.48 & 4.09 & 7.77 & 28.24 & 7.10 & 14.81 & 0.78 & 0.93 \\
            & HuatuoGPT & \textbf{27.39} & \textbf{14.38} & \textbf{8.06} & \textbf{4.55} & \textbf{8.52} & \textbf{29.26} & \textbf{8.02} & 15.46 & 0.74 & 0.93 \\
            \midrule
            \multirow{6}{*}{webMedQA} 
            & T5 (fine-tuned) & 21.42 & 13.79 & 10.06 & 7.38 & 8.94 & 31.00 & 13.85 & 25.78 & 0.37 & 0.46 \\
            % \hline
            \cline{2-12}
            & DoctorGLM & 9.91 & 5.20 & 2.78 & 1.54 & 4.67 & 23.01 & 5.68 & 11.96 & \textbf{0.84} & \textbf{0.95} \\
            & ChatGPT & 18.06 & 6.74 & 2.73 & 1.09 & 4.71 & 20.01 & 2.81 & 12.58 & 0.65 & 0.87 \\
            & ChatGLM-6B & 23.42 & 12.10 & 6.73 & 3.83 & \textbf{8.04} & \textbf{28.30} & 6.87 & \textbf{18.49} & 0.63 & 0.87 \\
            & Ziya-LLaMA-13B & 22.16 & 11.70 & 6.53 & 3.74 & 6.91 & 27.41 & 6.80 & 13.52 & 0.76 & 0.93 \\
            & HuatuoGPT & \textbf{24.85} & \textbf{13.42} & \textbf{7.72} & \textbf{4.51} & 7.50 & \textbf{28.30} & \textbf{7.72} & 14.50 & 0.73 & 0.93 \\
            \midrule
            \multirow{6}{*}{Huatuo-26M} 
            & T5 (fine-tuned) & 26.63 & 16.74 & 11.77 & 8.46 & 11.38 & 33.21 & 13.26 & 24.85 & 0.51 & 0.68 \\
            % \hline
            \cline{2-12}
            & DoctorGLM & 11.50 & 6.00 & 3.14 & 1.69 & 4.65 & 22.39 & 5.47 & 12.14 & \textbf{0.85} & \textbf{0.96} \\
            & ChatGPT & 18.44 & 6.95 & 2.87 & 1.13 & 4.87 & 19.60 & 2.82 & 12.46 & 0.69 & 0.89 \\
            & ChatGLM-6B & 24.46 & 12.75 & 7.20 & 4.13 & \textbf{8.50} & 28.44 & 7.31 & \textbf{18.58} & 0.67 & 0.89 \\
            & Ziya-LLaMA-13B & 25.58 & 13.39 & 7.46 & 4.24 & 7.30 & 28.14 & 7.18 & 14.78 & 0.77 & 0.93 \\
            & HuatuoGPT & \textbf{27.42} & \textbf{14.84} & \textbf{8.54} & \textbf{4.96} & 8.01 & \textbf{29.16} & \textbf{8.29} & 15.84 & 0.74 & 0.93 \\
            \bottomrule
        \end{tabular}
        \end{adjustbox}
    \end{table}

    \item \textbf{Law}: FMs can conduct in-depth analysis of legal documents and identify key information and legal concepts in the text, thereby assisting lawyers and legal advisors in more precise case analysis and legal consultation~\cite{lai2023large, zhou2024lawgpt, fei2024internlm}. For example, FMs can identify clauses in contracts, extract important legal elements such as obligations, rights, and conditions, help lawyers quickly understand document contents, and identify potential legal risks. Moreover, FMs can be used for case logical reasoning, predicting possible outcomes of cases by analyzing historical cases and relevant legal provisions and providing data support for lawyers to formulate defense strategies. The ChatLaw~\cite{cui2023chatlaw} can provide real-time legal consultation and answers, helping non-professionals understand complex legal issues and even generate drafts of legal documents, reducing the workload of lawyers. Additionally, FMs can assist in legal research, quickly retrieve relevant legal literature and precedents, and provide solid evidence for legal arguments.
    {However, LLMs can sometimes generate incorrect or fabricated information, known as ``AI hallucinations", which is a crucial challenge for domain-specific FMs in law. Another key challenge is the controversy about legal liability when using AI models for legal consultation. These challenges should be carefully addressed by practitioners in the process of developing legal models.}
    
    \item \textbf{Arts}: The application of FMs is exploring and changing the ways of creative expression and the process of artistic production. By learning from a large number of artworks and creative concepts, FMs can generate novel artworks, music, literary works, etc., providing creative inspiration and support for artists~\cite{shahriar2022gan}. For example, generative models can be used to produce artworks~\cite{schumacher2023enhancing, liang2024pianobart}, and fine-tuning techniques can be applied to adjust the style and content of the generated works~\cite{ho2020denoising, zhang2023adding}. Currently, in the field of video generation, Sora, developed by OpenAI, allows users to control generated content using text, producing lifelike video works. Furthermore, FMs also show promising performance in art understanding~\cite{zhao2024adversarial, paint_understand}, which provides great potential in the field of art teaching and study.
    {Acquiring training datasets in the art domain is highly contingent upon legal restrictions, particularly concerning copyright and intellectual property. Moreover, there is significant debate within artistic ethics about the appropriateness of employing AI models. These challenges are the key factors for FMs to contribute to art.}
    
    \item \textbf{Finance}: FMs can cover various tasks such as risk modeling, investment strategy management, market forecasting, etc., providing powerful tools for financial institutions and investors to optimize decision-making~\cite{li2023large, lopez2023can}. For example, FMs can be used to build credit scoring systems to assess borrowers' credit risks. These models analyze factors such as borrowers' historical credit records, financial conditions, and debt levels to predict their ability to repay loans and, based on this, decide whether to approve loan applications and the loan interest rates. Alternatively, FMs can be used to consider the historical performance, correlations, risks, and expected returns of various asset classes, as well as historical market data, macroeconomic indicators, political events, etc., to make optimal investor decisions. BloombergGPT, as detailed in~\cite{wu2023bloomberggpt}, was pre-trained on a large-scale financial dataset, resulting in superior performance across many benchmarks. \cite{yang2023fingpt} introduces FinGPT, an open-source financial LLM, and positions it not only as a model but also as an open-source framework for Financial LLMs (FinLLMs), which has the potential to fuel innovation among researchers.
    {Using time series analysis models to predict future stock trends and other financial information is common in the financial area. However, the accuracy of these models can significantly impact investors' decisions. Additionally, for LLMs to serve as effective investment advisors, they must possess more robust numerical analysis capabilities, which presents a crucial challenge in leveraging their full potential.}
\end{itemize}

% \section{Future directions}
\section{Challenges ahead}
The development of FM technology has achieved remarkable progress, but with the continuous advancement of technology, new challenges and issues arise, pointing out future research directions.

\subsection{Challenges in data}
\textbf{Acquiring domain-specific data and modeling data structures} are the top priority and most important challenges. FMs typically require vast amounts of high-quality data for training~\cite{du2024survey}, which may be costly and time-consuming to obtain in specific domains. Moreover, strengthening privacy regulations imposes more restrictions on data collection and usage. To address this challenge, future research can focus on developing new data collection and annotation techniques and utilizing synthetic data and weakly supervised learning methods to reduce reliance on extensively labeled data~\cite{zhou2018brief, van2020survey}. This reduces costs and effectively utilizes data resources while ensuring privacy protection. On the other hand, effective modeling of data structures in specific domains requires researchers and practitioners to deeply understand domain-specific business processes, extract key business data, and establish comprehensive data preprocessing pipelines.

\textbf{Understanding multi-modality data} poses another critical challenge. Although existing FMs excel in processing textual data, their understanding of other modalities, such as images and sounds, still needs improvement~\cite{zhang2024mm}, not to mention various new data modalities that may emerge in specific domains. Constructing unified models capable of comprehensively processing various modalities of data is essential for enhancing the model's performance and generalization on multi-modality tasks. This requires researchers and practitioners to not only deeply understand the characteristics of different modalities of data but also explore effective multi-modality fusion and interaction understanding mechanisms.

\textbf{The availability of aligned multi-modality data} presents a potential risk for the long-term development of multi-modal models. While techniques like bridging alignment can reduce the dependency on aligned data~\cite{girdhar2023imagebind, tang2024any}, the performance and generalization capabilities of multi-modal FMs still heavily rely on annotated data. This data is costly in terms of both time and financial resources. Therefore, new alignment methods are necessary to address these challenges~\cite{du2024probabilistic}.

\subsection{Challenges in model architecture}
In the architectural design of FMs in specific domains, a core challenge is how to build models that can \textbf{effectively capture and express deep semantic information} specific to different domain~\cite{girdhar2023imagebind, tang2024any, wu2023next}. This requires models to have a broad knowledge base and understand and adapt to domain-specific knowledge and input modalities. FMs tailored for specific domains need to achieve high \textbf{modularity and customizability} in architecture, enabling adjustment and optimization according to specific application scenarios to adapt to the data characteristics and task requirements of different domains~\cite{devlin2018bert, radford2018improving, raffel2020exploring}. Furthermore, \textbf{interpretability} of models is particularly important in specific domains. When designing architectures, researchers need to consider how to construct models so that their decision-making processes and output results can be understood and trusted by domain experts and end-users. This may involve developing new model mechanisms, introducing interpretable model intermediate representations, or designing visualization tools to demonstrate the internal workings of the model.

\subsection{{Challenges in model training and deployment cost}}
{
 The demands on computational, memory, and energy resources put significant limitations on the development of FM technologies. Therefore, their cost-intensive nature should be primarily considered when designing domain-specific FMs. Research on improving the efficiency of model training and inference~\cite{wang2024federated} and reducing resource consumption has become an urgent issue. Potential research directions include developing more efficient memory-saving and acceleration techniques, such as quantization, efficient attention, distributed learning, etc. These technologies can help model deployers reduce costs when maintaining model performance while building and deploying a domain-specific FM.
}

\subsubsection{{Challenges in cost-efficient training}}

{
The impressive capabilities of FMs come at a significant cost of computing, memory, and energy during training. For example, in some Internet of Things (IoT) scenarios, such costs could be the primary concern for model deployers. On one hand, they may need to use edge devices to collect real-time data for training large models rather than centrally pre-training a model and then deploying it. On the other hand, the resource demands of training models pose a hard constraint on edge devices. Since building domain-specific FMs is costly, it is beneficial to help model deployers better manage their costs.
}

{
% \textbf{Memory-saving techniques} can reduce iterations by processing more data in each iteration, thereby accelerating the training. 
\textbf{Memory efficiency:} FMs' huge parameter scale put a serious challenge for memory resources. Mixed precision training~\cite{micikevicius2017mixed} is a common technique to convert parts of high-precision variables into low-precision dynamically implemented by automatic mixed precision~\cite{lam2013automatically} in each training step. This can accelerate model training and reduce GPU memory usage while minimally impacting model accuracy. A huge memory is occupied while focusing on masked training in long-sequence training. Therefore, positional encoding compression methods like ALiBi~\cite{press2021train} and reset attention mask~\cite{dubey2024llama} have been proposed.
}

{
\textbf{Computation efficiency:} Most existing FMs have transformer-based architecture in which the attention mechanism has a powerful capability but is also computationally intensive. Researchers are now working to create efficient transformers. FlashAttention \cite{dao2022flashattention} significantly improves the efficiency of attention computation by IO-awareness block attention algorithm, and researchers also proposed other attention algorithms like LSH attention \cite{kitaev2020reformer} and local attention \cite{beltagy2020longformer}. Parallel training is another computational accelerating approach, including traditional data parallel and emerging methods like ZeRO method~\cite{rajbhandari2020zero} distribute data and model parameters to achieve fast training under limited hardware constraints. Additionally, parameter-efficient fine-tuning methods~\cite{houlsby2019parameter, hu2021lora} are also the key approach to improve the computation efficiency.
}

{
\textbf{Communication efficiency:} Federated learning~\cite{mcmahan2017communication} is applied when the training data cannot be acquired and centralized in advance due to privacy concerns. Reducing communication overhead is essential since FL involves frequent model parameter exchanges. Parameter-efficient fine-tuning methods have been combined with federated learning for efficient communication~\cite{sun2024improving, cho2024heterogeneous, zhang2023fedpetuning, kim2023client}.
}

{
\textbf{Energy efficiency:} The energy consumption of FM training is an emerging area as it generates significant carbon footprints~\cite{gupta2022act}, leading to environmental impacts at the same time as increasing economic costs.
}

\subsubsection{{Challenges in cost-efficient deployment}}

%Computational resources are also a significant challenge for FM technology. %Training and running FMs require enormous computational resources, which not only increases economic costs but also may have environmental impacts. 
{
The deployment of domain-specific FMs requires enormous resources, not only the training process but also the deployment of domain-specific FMs. In domain-specific business scenarios, the user's requirements for model response speed and performance may be an important consideration. To meet these demands under the resource constraints on the hardware, cost-efficient deployment strategies are highly desired, as introduced below:
}

{
\textbf{Inference efficiency:} The inference time of domain-specific FMs drastically increases with a huge amount of data. To solve such a problem, inference acceleration is crucial for deploying domain-specific FMs. KV cache~\cite{shi2024keep} is a popular method to speed up FM inference by only generating a single row of the Q matrix. Furthermore, multi-query attention~\cite{xu2023multi} could reduce the size of the KV cache, and paged attention~\cite{kwon2023efficient} utilizes the block table to make full use of the large KV cache. In addition to the improvement on transformer architecture, speculative decoding~\cite{leviathan2023fast} generates the next token with the help of a small “draft” model, which provides a set of token candidates for the main FM, followed by various methods such as staged speculative decoding~\cite{spector2023accelerating}, guided generation~\cite{willard2023efficient}, lookahead decoding~\cite{fu2024break} and prompt lookup decoding~\cite{saxena2023prompt}. Additionally, edge AI draws researchers' attention as it reduces the latency of providing AI services to devices~\cite{xing2023task, yang2020offloading, yang2021edge}. 
}

{
\textbf{Resource efficiency:} The resource requirements of FMs often make it difficult to deploy them in mobile devices and edge computational scenarios~\cite{yin2024llm}. To enable FMs to run in resource-constrained scenarios, lightweight model architectures and deployment strategies need to be developed. This may involve simplifying, distilling, and optimizing models to reduce their resource requirements while maintaining or improving their performance. Alternatively, employing cloud-edge collaboration strategies can distribute the training and inference processes of FMs across various server tiers, enabling a cooperative deployment. In cloud-edge learning, it is essential to consider the joint optimization of sensing, computation, and communication~\cite{xu2023edge, zhu2023pushing, wen2023task}. Such joint optimization for the performance of neural networks and resource consumption holds vast potential for innovation in research and can generate significant value in practical application scenarios. Furthermore, energy consumption in the inference phase is becoming the new trend within the community~\cite{ding2024sustainable}. There are trade-offs between performance and power usage to achieve energy-efficient FM inference~\cite{hisaharo2024optimizing, argerich2024measuring, liu2024energy, stojkovic2024towards}. Recently, edge AI inference~\cite{zhuang2023integrated, zhang2024collaborative} also shows potential to enable various devices without adequate computing resources to access the FMs.
}

\subsection{Challenges in security}
% Security issues are also challenges that FM technology must face. FMs may be used to generate false information, infringe on privacy, or be maliciously exploited, and the models themselves may also be subject to adversarial attacks~\cite{yao2024survey}. To ensure the security and reliability of models, relevant work includes but is not limited to the following key points: Firstly, \textbf{strengthening the robustness of models} to resist potential adversarial attacks~\cite{zou2023universal}. This may involve developing advanced adversarial training techniques and implementing more stringent data cleaning and preprocessing steps. Secondly, developing and deploying efficient \textbf{malicious input detection mechanisms}, using anomaly detection algorithms and real-time monitoring systems to identify and prevent malicious behavior. Furthermore, ensuring \textbf{privacy protection} is achieved by diminishing the model's reliance on sensitive data, thereby safeguarding the security and privacy of user information~\cite{feng2024exposing}. In addition to technical efforts, \textbf{enhancing security in workflow and policy aspects} is also necessary. For example, implementing security audit and certification processes to comprehensively assess the security of models and ensure that model development and deployment comply with ethical and legal standards, continuously updating security policies to address emerging security threats and challenges.

Security issues are also challenges that FM technology must face. FMs may be used to generate false information, infringe on privacy, or be maliciously exploited, and the models themselves may also be subject to adversarial attacks~\cite{yao2024survey}. To ensure the security and reliability of models, relevant work includes but is not limited to the following key points: Firstly, \textbf{strengthening the robustness of models} to resist potential adversarial attacks~\cite{zou2023universal}. This may involve developing advanced adversarial training techniques and implementing more stringent data cleaning and preprocessing steps. 
\textcolor{black}{In addition to traditional adversarial training, a representative line of early work proposed certified robustness methods—such as randomized smoothing~\cite{cohen2019certified}—which provide guarantees against perturbations. More recent studies have explored robust fine-tuning strategies, including adversarial contrastive learning and consistency regularization, to enhance generalization under distribution shifts. In parallel, robust representation learning has emerged as a promising direction, aiming to learn perturbation-invariant and semantically consistent embeddings, as exemplified by adversarial contrastive objectives in robust pre-training~\cite{jiang2020robust}. These complementary techniques contribute to a layered defense system that strengthens the reliability of foundation models in open environments.}
Secondly, developing and deploying efficient \textbf{malicious input detection mechanisms}, using anomaly detection algorithms and real-time monitoring systems to identify and prevent malicious behavior. Furthermore, ensuring \textbf{privacy protection} is achieved by diminishing the model's reliance on sensitive data, thereby safeguarding the security and privacy of user information~\cite{feng2024exposing}. In addition to technical efforts, \textbf{enhancing security in workflow and policy aspects} is also necessary. For example, implementing security audit and certification processes to comprehensively assess the security of models and ensure that model development and deployment comply with ethical and legal standards, continuously updating security policies to address emerging security threats and challenges.

\section{{Conclusion}}
{This paper provides a comprehensive overview of recent advancements in domain-specific foundation models (FMs). We begin by discussing the foundational concepts of FMs, including their architectures, training methodologies, the benefits of scaling, and a comparison of their performance across different contexts. Next, we delve into the key technologies for developing domain-specific FMs, which encompass three primary approaches: enhancing general-purpose FMs with domain-specific knowledge, customizing FMs using pre-trained modules, and building FMs from scratch without relying on pre-trained components. We then explore the diverse applications of domain-specific FMs across various fields, including telecommunications, autonomous driving, mathematics, medicine, law, the arts, and finance. Following this, we address the significant challenges faced by practitioners in constructing and deploying domain-specific FMs. These challenges span multiple dimensions, such as data availability and quality, model architecture design, training processes, deployment complexities, and security concerns. By synthesizing these insights, we aim to provide researchers and practitioners with a valuable resource for navigating the development and application of domain-specific FMs. We hope this paper serves as a foundation to guide future innovation and encourage the creation of tailored FMs that address specific domain requirements effectively.}

\Acknowledgements{This work was supported in part by National Natural Science Foundation of China (Grant No. 62371313), supported by the National Science and Technology Major Project (No. 2024ZD1300500), in part by Guangdong Major Project of  Basic and Applied Basic Research (No.  2023B0303000001), in part by Guangdong Young Talent Research Project (Grant No. 2023TQ07A708), in part by ShenzhenHong Kong-Macau Technology Research Programme (Type C) (Grant No. SGDX202308 21091559018), in part by the Shenzhen Science and Technology Program (Grant No. JCYJ20241202124934046), funded by Young Elite Scientists Sponsorship Program by CAST (2022QNRC001).}

\bibliographystyle{sciencechina}
\bibliography{reference}

% \bibliographystyle{plain}
% \usepackage{gbt7714}
% \bibliographystyle{gbt7714-numerical}
% \bibliography{reference}

%%%%%%%%%%%%%%%%%%%%%%%%%%%%%%%%%%%%%%%%%%%%%%%%%%%%%%%
%%% Appendix sections. ¸½Â¼ÕÂ½Ú, ·Ç±ØÑ¡
%%%%%%%%%%%%%%%%%%%%%%%%%%%%%%%%%%%%%%%%%%%%%%%%%%%%%%%
%\begin{appendix}
%\section{Name}

%\end{appendix}

\end{document}